\documentclass[sn-basic]{sn-jnl}

\usepackage{graphicx}%
\usepackage{multirow}%
\usepackage{amsmath,amssymb,amsfonts}%
\usepackage{amsthm}%
\usepackage{mathrsfs}%
\usepackage[title]{appendix}%
\usepackage{xcolor}%
\usepackage{textcomp}%
\usepackage{manyfoot}%
\usepackage{booktabs}%
\usepackage{algorithm}%
\usepackage{algorithmicx}%
\usepackage{algpseudocode}%
\usepackage{listings}%
\usepackage{ragged2e}
\usepackage{adjustbox}
\usepackage{tabularx}
\usepackage{subcaption}

\theoremstyle{thmstyleone}%
\theoremstyle{thmstyletwo}%

\theoremstyle{thmstylethree}%

\begin{document}

\title[Unified Taxonomy for MTSAD using DL]{Unified Taxonomy for Multivariate Time Series Anomaly Detection using Deep Learning}


\author*[1]{\fnm{Bruna} \sur{Alves}}\email{bruna.alves@ua.pt}

\author[1]{\fnm{Armando J.} \sur{Pinho}}\email{ap@ua.pt}

\author*[1]{\fnm{Sónia} \sur{Gouveia}}\email{sonia.gouveia@ua.pt}

\affil*[1]{\orgdiv{IEETA/DETI/LASI}, \orgname{University of Aveiro}, \orgaddress{\postcode{3810-193}, \state{Aveiro}, \country{Portugal}}}


\abstract{The topic of Multivariate Time Series Anomaly Detection (MTSAD) has grown rapidly over the past years, with a steady rise in publications and Deep Learning (DL) models becoming the dominant paradigm. To address the lack of systematization in the field, this study introduces a novel and unified taxonomy with eleven dimensions over three parts (Input, Output and Model) for the categorization of DL-based MTSAD methods. 
The dimensions were established in a two-fold approach. First, they derived from a comprehensive analysis of methodological studies. Second, insights from review papers were incorporated. Furthermore, the proposed taxonomy was validated using an additional set of recent publications, providing a clear overview of methodological trends in MTSAD.
Results reveal a convergence toward Transformer-based and reconstruction models, setting the foundation for emerging adaptive and generative trends. Building on and complementing existing surveys, this unified taxonomy is designed to accommodate future developments, allowing for new categories or dimensions to be added as the field progresses. This work thus consolidates fragmented knowledge in the field and provides a reference point for future research in MTSAD.}


\maketitle

\section{Introduction}

The rapid advancement of digital technologies and widespread connectivity has led to a significant increase in the deployment of sensors across a wide range of domains. These sensors continuously monitor their surrounding environments, generating large volumes of data - often as Multivariate Time Series (MTS), where multiple variables are recorded simultaneously over time. A key goal in analyzing these data is Anomaly Detection (AD), which refers to the identification of deviations from expected or typical patterns observed within the data \citep{Audibert2022, Chandola2009}.

Multivariate Time Series Anomaly Detection (MTSAD) has gained relevant attention in recent years, reflected by a growing interest of the scientific community and its widespread adoption across domains such as industrial systems \citep{WADI}, server monitoring \citep{OmniAnomaly} and spacecraft operations \citep{MSL}. 
Unlike Univariate Time Series (UTS) anomaly
detection methods \citep{BlzquezGarca2021}, which model only temporal
dependencies within a single series, MTSAD requires capturing both temporal and inter-metric (i.e., spatial) dependencies that reflect relationships among variables. This results in a more complex modeling task, as methods must account for both intra- and inter-series dynamics. When such multivariate dependencies are present, this spatio-temporal setting allows the identification of a broader range of anomaly patterns that may affect one, several, or all variables
simultaneously \citep{Fu2024, Guan2024, He2020, Tian2023,  MSCRED, MTAD-GAT}.
Given the complexity of MTSAD task, conventional statistical methods like Vector Autoregressive (VAR) models and classical Machine Learning (ML) approaches, such as K-means clustering and Isolation Forest (IF) \citep{Audibert2022} originally designed for UTS, fall short in addressing its challenges. In contrast, Deep Learning (DL) models have emerged as state-of-the-art solutions, due to their ability to model non-linear temporal dynamics and dependencies across multiple variables that traditional methods often miss.

Reflecting the rapid evolution of the research field, several review papers on MTSAD have been published recently \citep{Belay2023, Choi2021, Correia2024, ElAmineSehili2024, Fahrmann2024, Li2023, Orabi2024, Pea2023, Ramezani2021, Usmani2024, ZamanzadehDarban2024}. Some focus on specific topics like evaluation metrics \citep{ElAmineSehili2024}, Hidden Markov Models \citep{Ramezani2021} or statistical outlier detection \citep{Pea2023}, while others offer broader surveys on both ML and DL approaches \citep{Belay2023, Fahrmann2024, Orabi2024, Usmani2024}. Some of the reviews propose categorizations specifically for core DL architectures based on Autoencoders (AE), Variational Autoencoders (VAE), Recurrent Neural Networks (RNN), Transformers, Hierarchical Temporal Memory (HTM), Temporal Convolutional Networks (TCN), Graph Neural Networks (GNN) and Generative Adversarial Networks (GAN) \citep{Belay2023, Choi2021, Fahrmann2024, Usmani2024, ZamanzadehDarban2024}. In addition, certain reviews explore AD strategies, typically grouped into reconstruction-based, prediction-based and other variants \citep{Belay2023, Choi2021, Correia2024, ZamanzadehDarban2024}. A number of reviews also classify methods based on the type of detected anomaly (e.g. time point, time interval or full time series) \citep{Correia2024, Li2023, ZamanzadehDarban2024}, though these categorizations often lack consistency across authors.

An important dimension to consider is how DL models capture temporal and spatial dependencies. For instance, \citet{Choi2021} emphasize this distinction and \citet{ZamanzadehDarban2024} categorize methods as Temporal (T), Spatial (S) or Spatio-temporal (ST). \citet{Li2023} further discuss thresholding techniques, feature extraction and evaluation metrics. Despite these valuable contributions, the DL-based MTSAD field still lacks a unified framework. While several reviews cover similar dimensions, they often propose independent and sometimes inconsistent categorizations. Moreover, key aspects remain underexplored, e.g. how spatial and temporal dependencies are jointly modeled in spatio-temporal approaches. 

To address these gaps, this paper proposes a unified taxonomy that, unlike existing surveys, systematically covers the key aspects of DL-based models targeting MTSAD. Built on concepts from recent reviews \citep{Choi2021,Correia2024,Li2023,ZamanzadehDarban2024} and enriched by insights from an extensive screening of literature published between 2019 and 2024 indexed at Scopus\textregistered\ (\href{https://www.scopus.com/}{https://www.scopus.com/}), it comprises eleven key dimensions grouped into three main parts characterizing Input, Output and Model. This provides a consistent and extensive framework that enables direct comparison across methods and explicitly addresses aspects that have so far remained underexplored, highlighting opportunities for future research. This novel taxonomy aims to serve as a practical guide for researchers and practitioners, as well as a starting point for newcomers to the field. In the near future, the taxonomy is expected to be expanded and refined alongside advances in model architectures, detection strategies and data representations.

This paper is organized in a crescendo of detail and complexity. Section~\ref{sec2} serves as a 
background for readers entering the research field, providing key concepts, problem 
formulation, performance evaluation methodologies, and benchmarks for MTSAD. 
Section~\ref{sec3} describes the methodology used to construct the taxonomy, starting 
with the paper screening and selection process and culminating in the proposed taxonomy 
of DL-based approaches for MTSAD, complemented by insights gathered from previous 
reviews. Section~\ref{sec4} presents a visual summary of the literature distribution 
across the taxonomy dimensions and discusses representative approaches within each 
dimension, highlighting their design choices, strengths, and limitations. 
Section~\ref{sec5} examines the current research paradigm, identifies prevailing 
patterns, and outlines recent developments, ongoing challenges, and future research 
directions. Finally, Section~\ref{sec6} concludes the paper.

\section{Concepts, Methods and Benchmarks in MTSAD}\label{sec2}

In general terms, an anomaly is an observation (or a set of consecutive temporal observations) that largely deviates from what is considered the normal behavior, which may indicate abnormal or problematic system behavior \citep{Audibert2022, Chandola2009}. In practice, it is first necessary to define what the ``normal'' behavior is — typically by relying on a model trained with examples of historical data previously labelled as normal by a domain expert, as it will be discussed shortly. The new data is then flagged as an anomaly when it deviates from the expected range, pattern or structure given the model for the ``normal'' behavior. It is important to note that anomalies are often mistaken for novelties or outliers and while the concepts may overlap, they also hold fundamental differences. On one hand, all three concepts concern ``unusual'' data that deviates from what has been defined as the normal pattern. On the other hand, they differ in intent and interpretation: novelties are previously unseen but potentially normal observations and outliers are typically statistically rare events, both not necessarily implying an abnormal system behavior \citep{Pimentel2014}.

\subsection{Types of anomalies: temporal versus inter-metric}

\begin{figure}[h!]
    \centering
    \includegraphics[width=0.9\linewidth]{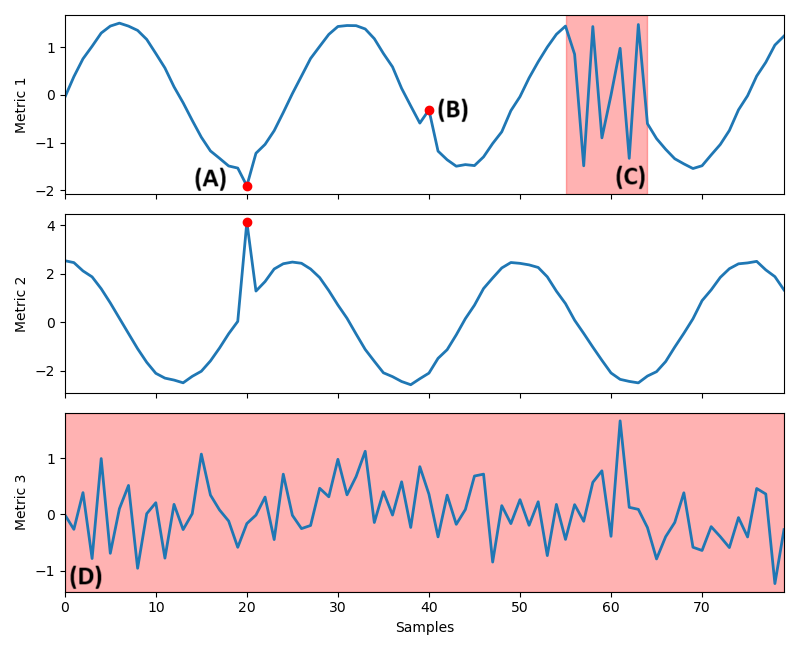}
    \caption{
    Examples of temporal anomalies in synthetic MTS generated with \url{https://github.com/datamllab/tods/tree/benchmark/benchmark/synthetic/Generator}: (A) Point anomaly in metrics 1 and 2, (B) Contextual in metric 1, (C) Collective in metric 1 and (D) Collective spanning metric 3. The anomaly in (C) is also inter-metric.}
    \label{fig:anomaly_types}
\end{figure}

When dealing with MTS data, anomalies are likely to arise along the temporal dimension (temporal anomaly) and across different metrics at a given time point (inter-metric anomaly). While a temporal anomaly is typically defined along the temporal axis for each metric (e.g. \textit{Point}, \textit{Contextual} or \textit{Collective} anomaly) \citep{Chandola2009}, the inter-metric anomaly occurs when the relationship between two or more metrics changes within a specific time interval, even if the individual metrics appear normal when considered separated \citep{Interfusion}. In practice, a temporal anomaly in a subset of metrics often corresponds to a temporal inter-metric anomaly, since deviations in one metric may affect the correlation with other metrics. Therefore, anomalies in MTS can manifest both temporally and inter-metrically, which require detection methods that capture the temporal dynamics along each metric as well as the relationship across metrics. 
Figure~\ref{fig:anomaly_types} shows representative examples of anomalies in MTS data. A temporal anomaly typically manifests when the evolution of a single metric deviates from its historical or expected temporal pattern, such as an abrupt spike (\textit{Point} anomaly, A), an unexpected deviation in a given context (\textit{Contextual} anomaly, B) or a subsequence of values that collectively exhibit abnormal behavior (\textit{Collective} anomaly, C and D) \citep{Chandola2009}. In contrast, an inter-metric anomaly occurs when the statistical or functional relationship between two or more metrics changes, even if the individual metrics appear normal in isolation \citep{Interfusion}. This is the e.g. case for the anomaly (C) in Fig.~\ref{fig:anomaly_types} which is also inter-metric as it disrupts the dependency structure between metrics. 

Several taxonomies have been proposed to categorize anomalies in time series, ranging from the classical framework of \citet{Chandola2009} proposed and adopted for UTS \citep{Audibert2022, Usmani2024, Jia2025} to more recent schemes addressing multivariate and inter-metric dependencies \citep{Interfusion}. \citet{Chandola2009} proposed the above-mentioned categories, namely \textit{Point}, \textit{Contextual} and \textit{Collective} for temporal anomalies. \citet{Lai2021} refined this taxonomy by subdividing Point anomaly into \textit{Global} or \textit{Local} and \textit{Pattern} anomaly (equivalent to Collective anomaly of \citet{Chandola2009} into \textit{Shapelet}, \textit{Seasonal} or \textit{Trend}.

The extension for MTS was introduced by \citet{Interfusion}, with categories of \textit{Temporal}, \textit{Inter-metric} and \textit{Temporal–inter-metric} anomaly. \citet{ZamanzadehDarban2024} integrated the temporal categories of \citet{Lai2021} with the multivariate types of \citet{Interfusion}. \citet{Wang2025} further refined Inter-metric anomalies into \textit{Global–inter-metric} (entire metric affected) and \textit{Local–inter-metric} (correlations disrupted over intervals). As a final note, other surveys classify anomalies by length as \textit{Point}, \textit{Sub-sequence} or \textit{Sequence} (entire time series) anomalies \citep{BlzquezGarca2021, Correia2024, Li2023}.

\subsection{Problem Formulation in MTSAD}
The objective of MTSAD is to detect abnormal patterns while capturing both temporal dynamics and inter-metric dependencies. Formally, the task is to learn a model $f$ that maps the input data $\mathbf{X}$ into a known output $y$
\begin{equation}
f: \mathbf{X} \rightarrow y \in \{0,1\}, 
\end{equation}
where typically $y=1$ is used to identify anomaly and $y=0$ for non-anomaly.

The input data $\mathbf{X}$ can take different formats. In the simplest case, $\mathbf{X} := \mathbf{x}_{\bullet,t} \in \mathbb{R}^{M \times 1}$ is the vector 
\begin{equation}
\mathbf{x}_{\bullet,t}=\begin{bmatrix}  {x}_{1,t} & {x}_{2,t} & {x}_{3,t} & \cdots & {x}_{M,t}\end{bmatrix}^{\top} 
\end{equation}
that represents the information at time point $t=1, \cdots, T$ over all $m= 1, \cdots, M$ metrics \citep{Rancoders}. Alternatively, $\mathbf{X} := \textbf{x}_{\bullet,t:t+w} \in \mathbb{R}^{M \times (w+1)} $ can be a subsequence (or window) of length $w+1$ denoted as

\begin{equation}
\textbf{x}_{\bullet,t:t+w}= \begin{bmatrix}
\mathbf{x}_{\bullet,t} &\mathbf{x}_{\bullet,t+1} & \cdots & \mathbf{x}_{\bullet,t+w}    
\end{bmatrix}=
\begin{bmatrix}
{x}_{1,t} & {x}_{1,t+1} &  \cdots & {x}_{1,t+w} \\
{x}_{2,t} & {x}_{2,t+1}  & \cdots & {x}_{2,t+w} \\
\vdots & \vdots &  \ddots & \vdots \\
{x}_{M,t} & {x}_{M,t+1} & \cdots & {x}_{M,t+w}
\end{bmatrix} ,
\label{eq:window}
\end{equation}
which captures both short-term temporal patterns and associations across metrics \citep{OmniAnomaly, USAD}. This unified notation allows the framework to detect \textit{Contextual} and \textit{Collective} anomalies besides \textit{Point} ones, via window-based detection. In certain settings, the raw time series may be further transformed into alternative representations, such as descriptive measures based on sufficient statistics, learned embeddings or even images, that can also serve as input to the model \citep{MSCRED}.

In classification settings, the model directly provides a binary output $\hat{y} = f(\mathbf{X})$ given the input $\mathbf{X}$ \citep{MBead}. On the other hand, a regression model produces an output $\hat{\mathbf{X}} = f(\mathbf{X})$, which can be a representation (embedding) \citep{DCdetector}, reconstruction \citep{OmniAnomaly} or prediction \citep{GDN} based on the input data. The anomaly score is then computed as the dissimilarity between the observed $\mathbf{X}$ and the model output $\hat{\mathbf{X}}$,
\begin{equation}\label{eq: AS}
AS = diss(\mathbf{X}, \hat{\mathbf{X}}).
\end{equation}
Thus, while the regression output is an approximation of $\mathbf{X}$ against which deviations are measured, the classification output directly represents the anomaly label of $\mathbf{X}$. It is worth noting that in probabilistic classification, the model predicts $P(y=1|\mathbf{X}) \in [0,1]$, which can be interpreted as a continuous anomaly score, making it conceptually close to the regression setting. In both cases, a higher probability or dissimilarity score $AS$ indicates a greater likelihood of anomaly. A thresholding rule is then applied to $AS$ to obtain the binary output $\hat{y}$ as

\begin{equation}\label{eq: threshold}
  \hat{y}=g(AS)=\begin{cases}
    1, & \text{if $AS > \delta$}\\
    0, & \text{otherwise}
  \end{cases} \quad ,
\end{equation}
where $\delta \in \mathbb{R}^+$ denotes the decision threshold.

The dissimilarity $AS$ is typically computed using a distance or similarity measure, such as correlation or cosine similarity \citep{Levy2025}. Regarding distances, a common choice is the $L_p$-norm defined as
\begin{equation}
\| \textbf{v} \|_p = \left( \sum_{i=1}^{n} |v_i|^p \right)^{\frac{1}{p}},
\end{equation}
where $\mathbf{v}=\begin{bmatrix} v_1 & v_2 & \cdots &v_n    
\end{bmatrix} \in \mathbb{R}^{1\times n}$ represents a vector of differences between observed and model output. Different well-known distances arise from this class, namely the Manhattan ($p=1$), Euclidean ($p=2$) and Chebyshev ($p \to \infty$) distance. This concept extends to matrices via entry-wise norms, 
\begin{equation}
\| \mathbf{A} \|_{(p,q)} = \left( \sum_{i=1}^{k} \left( \sum_{j=1}^{n} |a_{ij}|^q \right)^{p/q} \right)^{1/p},
\end{equation}
with $\mathbf{A}= \begin{bmatrix} a_{ij}
\end{bmatrix}  \in \mathbb{R}^{k\times n}$, $i=1, \cdots,k$ and $j= 1, \cdots, n$. This includes the Frobenius norm ($p=q=2$), often treated as the Euclidean distance of a matrix. Note that in the literature, the notation $\|\cdot\|_2$ can be used ambiguously for both the Euclidean vector norm and the Frobenius matrix norm.

\begin{figure}[h!]
    \centering
    \includegraphics[width=0.6\linewidth]{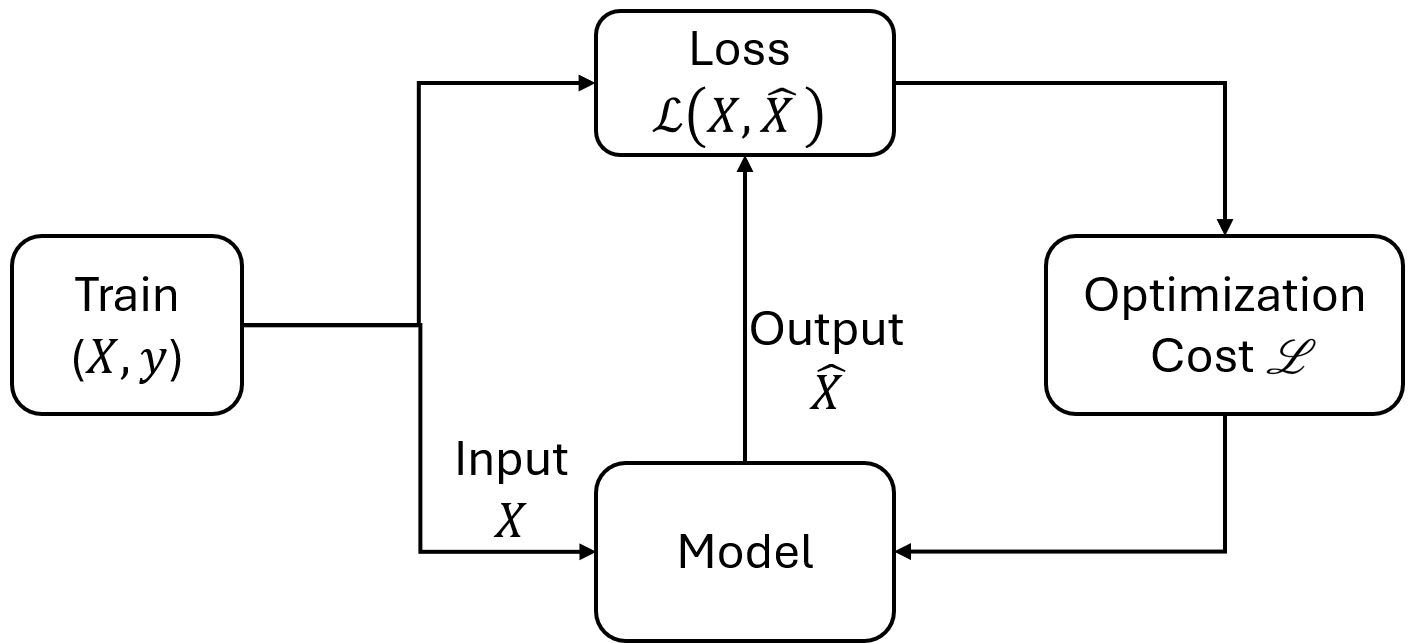}
    \caption{Schematic representation of the process of learning a regression model $f$. In case of a classification model, the output of the model would be $\hat{y}$ and the loss function would compare it with the know target $y$, $\mathcal{L}(y,\hat{y})$.}
    \label{fig:framework}
\end{figure}

Figure~\ref{fig:framework} illustrates the learning process of the regression model $f$ for MTSAD. The data input $\mathbf{X}$ is fed through the model to produce an output which is then compared to the corresponding target using a loss function $\mathcal{L}$ that quantifies the error in the model output. The overall error of the model is computed through a cost function $\mathscr{L}$ that aggregates the loss values over $N$ different inputs as 

\begin{equation}
    \mathscr{L}= \frac{1}{N}\sum_{i=1}^N\mathcal{L}_{i}.
    \label{eq:cost_function}
\end{equation}

\noindent  This error is backpropagated through the network, enabling the model to update its weights via an optimization algorithm, typically gradient descent. Training is usually performed on a subset of the dataset chosen for that purpose, referred to as the training set. During training, a separate validation set is used to monitor the generalization performance of the model, helping to detect and prevent overfitting. It also guides hyperparameter tuning, often via techniques such as cross-validation \citep{Duda_and_Hart}. The test set is then reserved for the final performance evaluation of the model after training and hyperparameter selection. For regression-based anomaly detection, the training set typically consists of examples representing normal behavior (i.e., $y=0$). Consequently, the model output $\hat{\mathbf{X}}$ is expected to closely match normal patterns. On the other hand, classification models are trained on labeled examples of anomaly ($y=1$) and non-anomaly ($y=0$) and the model predicts a label $\hat{y} \in \{0,1\}$ for each input or optionally a probability $P(y=1 \mid \mathbf{X})$.

\subsection{Evaluation of an MTSAD methodology}\label{sec: performance_evaluation}

The performance of AD methods can be assessed using non-binary or binary metrics \citep{Evaluation}. Non-binary metrics assess model performance across varying thresholds, whereas binary metrics evaluate the predictions after applying a specific threshold value, $\delta$. The most commonly used non-binary metrics are the Area Under the Receiver Operating Characteristic curve (AUROC) and the Area Under the Precision-Recall curve (AUPRC) \citep{Evaluation}. Briefly, AUROC summarizes the trade-off between the True Positive Rate (TPR) and False Positive Rate (FPR) across different detection thresholds, 
\begin{equation}
TPR = \frac{TP}{TP + FN},
\end{equation}
\begin{equation}
FPR = \frac{FP}{FP + TN},
\end{equation}
where True Positives ($TP$), False Negatives ($FN$), False Positives ($FP$) and True Negatives ($TN$) are defined with anomalies being the positive class.
On the other hand, AUPRC summarizes Precision versus Recall across thresholds, where
\begin{equation}
Precision = \frac{TP}{TP + FP}
\end{equation}
is the proportion of correctly identified anomalies among all predicted anomalies and \textit{Recall} (equivalent to $TPR$) is the proportion of correctly identified anomalies among all true anomalies. In a comparison between these metrics, AUPRC is generally considered more informative for imbalanced problems, which is the typical setting in anomaly detection \citep{Evaluation}. In this case, the large $TN$ keeps the $FPR$ low even with many $FP$, making AUROC less sensitive to errors and less reflective of the ability of the model to detect rare anomalies \citep{Evaluation}.

All the above-mentioned metrics can be used to evaluate the performance of the model at that particular decision threshold $\delta$. A popular choice in MTSAD problems is the F1-score,
\begin{equation}
F_1=2 \times \frac{Precision \times Recall}{Precision+Recall},
\end{equation}
defined as the harmonic mean of \textit{Precision} and \textit{Recall}. This measure is particularly suitable for imbalanced datasets as it reflects the ability of the model to detect rare anomalies without being dominated by the large number of non-anomaly samples.

\begin{figure}[h]
    \centering
    \includegraphics[width=0.9\linewidth]{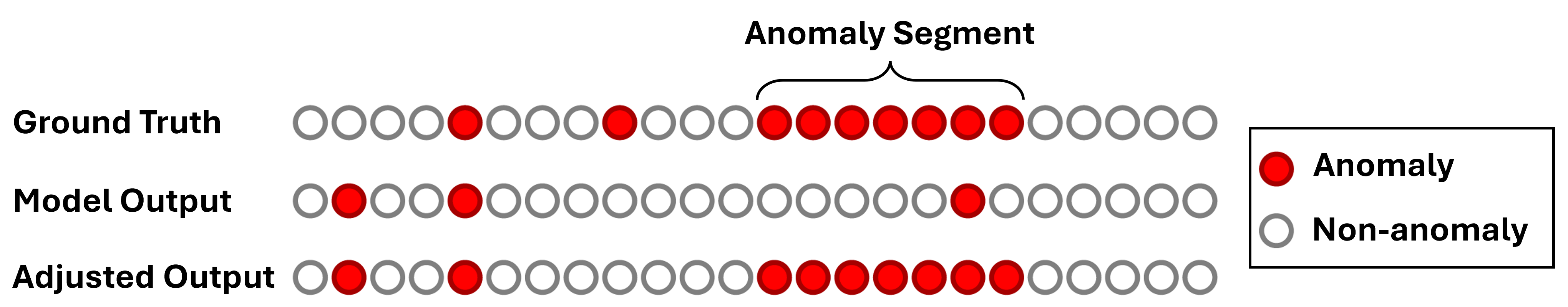}
    \caption{Illustration of the point-adjustment (PA) strategy showing ground truth, model output and adjusted output over time points (circles). Image adapted from \citep{ElAmineSehili2024}.}
    \label{fig: pa}
\end{figure}

MTSAD evaluation is usually based on binary metrics computed using the point-adjustment (PA) strategy \citep{USAD, OmniAnomaly, MTAD-GAT, Interfusion, GTA}. As illustrated in Fig.~\ref{fig: pa}, PA considers an entire anomaly segment as correctly detected if at least one point of that segment is flagged, treating all points in the segment as True Positives, while points outside segments are evaluated as non-anomaly. This reflects the fact that anomalies often occur over continuous periods and that an alert anywhere in a segment is typically sufficient \citep{pa}. However, PA can overestimate performance and assume homogeneity within segments \citep{Kim2022, ElAmineSehili2024}. To address this, \citet{Kim2022} proposed PA\%K, which requires a minimum fraction $K$ of point anomalies in a segment to be detected, providing a fairer assessment.

\subsection{Benchmark MTSAD methodologies and datasets}

MTSAD methods have applications across a diversity of real-world domains, including industrial systems such as power plants \citep{MSCRED, Choi2020}, IoT and cyber-physical systems such as water distribution networks \citep{GDN, MAD-GAN}, robotics \citep{SWCVAE, LSTM-VAE}, server machine monitoring \citep{OmniAnomaly, Rancoders}, healthcare \citep{CAE-M}, space telemetry \citep{MSL} and structural health monitoring \citep{STVAE}.

Comparing MTSAD methodologies is challenging due to variations in datasets, evaluation metrics and experimental setups. Several real-world data benchmarks have been widely adopted, including SMAP and MSL \citep{MSL}, SWaT \citep{SWAT} and WADI \citep{WADI} and SMD \citep{OmniAnomaly} and PSM \citep{Rancoders}. Synthetic datasets, such as NeurIPS-TS \citep{Lai2021}, offer controlled anomaly rates and types, facilitating systematic evaluation of model behavior under different conditions. See other examples of data benchmarks and their domains in Table 3 of \citep{ZamanzadehDarban2024}.

Several studies have systematically compared state-of-the-art MTSAD methods under controlled conditions \citep{Audibert2022, Choi2021, Garg2022, Li2023b, Schmidl2022}. For instance, \citet{Choi2021} evaluated ten DL-based methods on SWaT, WADI and MSL datasets, reporting the hyperparameters (e.g. window size and learning rate) for reproducibility. In addition, benchmarking toolkits such as TimeEval \citep{WenigEtAl2022TimeEval}, TODS \citep{Lai_Zha_Wang_Xu_Zhao_Kumar_Chen_Zumkhawaka_Wan_Martinez_Hu_2021} and TimeSeAD \citep{timesead} have been developed to facilitate standardized comparisons. Certain models have also become widely adopted benchmarks, including conventional and ML-based approaches like Principal Components Analysis, Matrix Profile, IF and One-class Support Vector Machine (OCSVM) \citep{Audibert2022, Schmidl2022}, as well as DL-based models such as DAGMM \citep{DAGMM}, LSTM-VAE \citep{LSTM-VAE}, OmniAnomaly \citep{OmniAnomaly}, MAD-GAN \citep{MAD-GAN}, MSCRED \citep{MSCRED}, THOC \citep{THOC}, USAD \citep{USAD}, GDN \citep{GDN} and GTA \citep{GTA}, which are commonly used as baselines in comparative studies.

\section{Taxonomy of DL-based approaches for MTSAD}\label{sec3}

This paper proposes a taxonomy for DL in MTSAD to provide a structured organization of the state-of-the-art, unifying the results from the analysis of 409 documents and 12 reviews or evaluation papers. The full list of categorized papers and reviews is available in the supplementary material of this paper. Figure \ref{fig:workflow} illustrates the paper screening process, resulting in 409 categorized papers from an initial pool of 898 screened documents. The following sections present the paper screening process and selection criteria, publication trends in the research field and the proposed taxonomy in detail.

\begin{figure}[h!]
    \centering
    \includegraphics[width=0.8\linewidth]{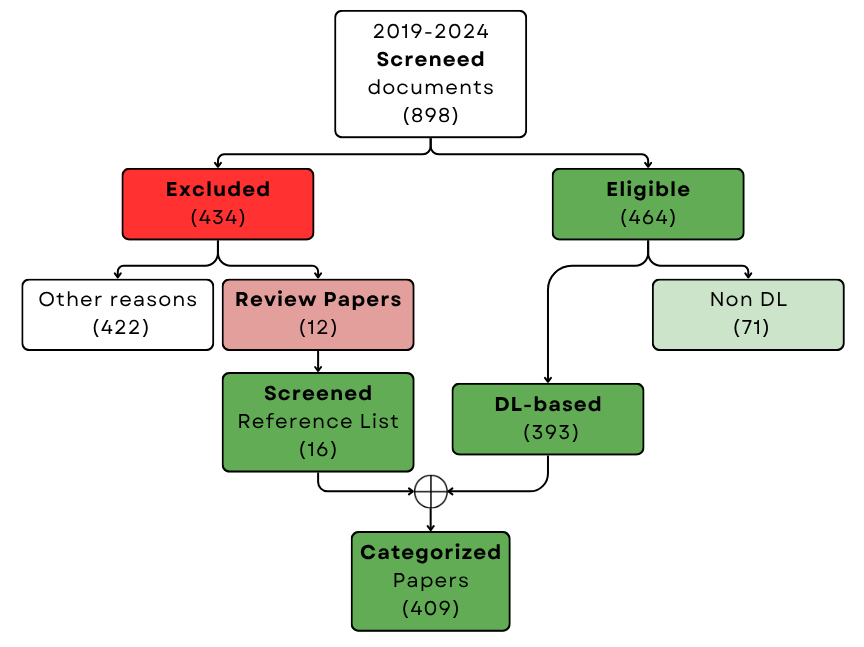}
    \caption{Paper screening and selection process. Starting from 898 screened documents (2019--2024), 393 DL-based papers were selected and combined with 16 papers from reference list screening, resulting in 409 categorized papers. Additionally, 12 review papers were identified to complement the proposed taxonomy.}
    \label{fig:workflow}
\end{figure}

\subsection{Paper screening and selection criteria} 

The screening process followed the PRISMA (Preferred Reporting Items for Systematic reviews and Meta-Analyses) guidelines, covering the stages of identification, screening, eligibility and inclusion/exclusion criteria \citep{Moher2009}. The relevant literature was identified through an automatic search for documents in one of the most extensive abstract and citation databases of peer-reviewed literature (Scopus \textregistered, \href{https://www.scopus.com/}{https://www.scopus.com/}) followed by a manual screening process based on the content of each document. The search procedure was based on the following advanced query string\\

\setlength{\parindent}{20pt}
\parbox{\textwidth}{\ttfamily \justifying
\noindent TITLE-ABS-KEY("multivariate time series" AND \\
("anomaly detection" OR \\
"novelty detection" OR "outlier detection" OR \\
"pattern recognition")) \\ 
AND PUBYEAR > 2018 AND PUBYEAR < 2025 AND LANGUAGE(english)}
\setlength{\parindent}{15pt}\\

\noindent to return all documents that included specific search terms (in the title, abstract or keywords field), were published between January 1, 2019 and December 31, 2024, and were written in English. 
The search was last updated on June 30, 2025, to ensure the inclusion of all publications from 2024 that matched the criteria, as some papers indexed in 2024 were added to the database with a delay.

The search term was defined as the intersection of "multivariate time series" and terms related with AD methods. Since "anomaly detection", "novelty detection" (ND) and "outlier detection" (OD) are frequently used interchangeably in the literature to refer to the identification of unusual patterns in data, all three terms were included in the search. Additionally, the broader term "pattern recognition" was considered to include documents that may propose new methodologies, which can overlap with anomaly, novelty or outlier detection, even if these terms are not specifically addressed in the title, abstract or keywords field. The search covered all types of publications, including papers and reviews, available in both journal and conference proceedings formats. No methodological terms (e.g. "deep learning") were included to avoid introducing bias towards specific approaches, thus ensuring that the query collected all the relevant literature for this review.

The search results were exported into a spreadsheet and the publications were ranked by decreasing citation count to identify the most cited references while eliminating possible duplicate contributions. The screening process focused on including publications proposing new methodologies for MTSAD. All documents underwent a manual screening process in two consecutive steps: title/abstract evaluation and full-text reading. In the first step, two evaluators analyzed each title/abstract, and publications were excluded if both agreed they did not meet the inclusion criterion. In the second step, the selected documents followed to full-text reading and those meeting the criteria were included. Excluded publications fell into several categories, such as
\begin{enumerate}
    \item \textbf{review articles};
    \item \textbf{other contributions} (e.g. explainability methods, tools, datasets, input transformations, handling of missing values and decision rules) or \textbf{duplicated} contributions;
    \item \textbf{applications} studies that evaluate specific methods or compare multiple approaches using either benchmark datasets or non-public datasets;
    \item publications \textbf{not addressing AD, ND or OD} methodologies explicitly;
    \item \textbf{errata}, conference or challenges \textbf{overviews};   
    \item \textbf{not open access}, either freely online, or via the Scopus institutional partnership with the University of Aveiro or the University of Porto, Portugal. Articles not available on Scopus were requested via ResearchGate (\href{https://www.researchgate.net/}{https://www.researchgate.net/}) when possible.
\end{enumerate}

After the screening process, the eligible documents were then categorized into three classes, namely conventional, ML- and DL-based approaches, following the classification framework of \citet{Audibert2022}. Briefly, according to these authors, conventional approaches estimate model parameters based on the assumption that data is generated from a stochastic process, typically linear and Gaussian. ML-based models, on the other hand, predict outcomes by learning patterns in data without a predefined model, while DL-based approaches refer to models that consist of multiple stacked layers.

As previously mentioned, review papers were excluded from the core set of publications analyzed in this review; however, they were used in two complementary ways. First, they served to support the empirical development of the proposed taxonomy. Second, their reference lists were screened to identify additional relevant papers that might have been missed by the initial query. Specifically, references with a high number of citations (i.e., more than 25 citations/year since publication), published between 2019 and 2024 and presenting clear methodological contributions to MTSAD using DL-based approaches, were included in this review.

\subsection{Publication trends in MTSAD}

\begin{figure}
    \centering
    \includegraphics[width=\linewidth]{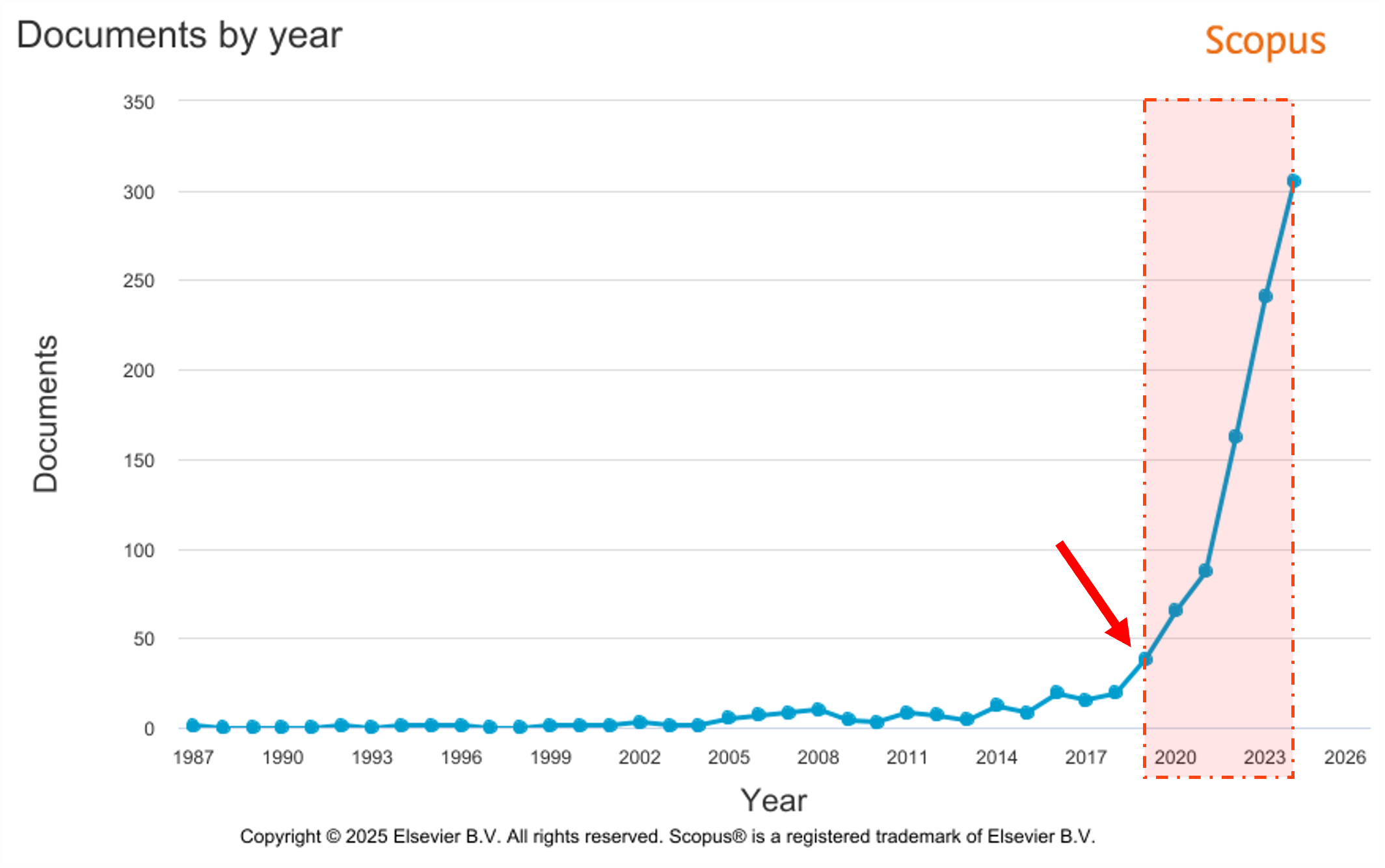}
    \caption{Annual number of documents retrieved using the search query. A total of 1,040 documents were retrieved, including 142 before 2019 and 898 from 2019 to 2024. The red-shaded area highlights the 2019–2024 period, which marks a significant growth in publication volume. Chart directly retrieved and adapted from Scopus \textregistered (\href{https://www.scopus.com/}{https://www.scopus.com/}).}
    \label{fig:trend}
\end{figure}

A preliminary analysis of the publication trends was conducted to contextualize the evolution of research activity in the field. Figure~\ref{fig:trend} presents the number of documents retrieved from Scopus over the years using the defined search query, without applying any time interval restrictions. A total of 1,041 documents were retrieved. A clear turning point in the research landscape appears to be the year 2019, with only 143 documents before, compared to 898 documents published during the 2019–2024 period. The number of publications has steadily increased each year since 2019, reflecting growing interest and accelerated research activity in the field. For this reason, the present review focuses specifically on the period from 2019 to 2024.

\begin{figure}
    \centering
    \includegraphics[width=0.8\linewidth]{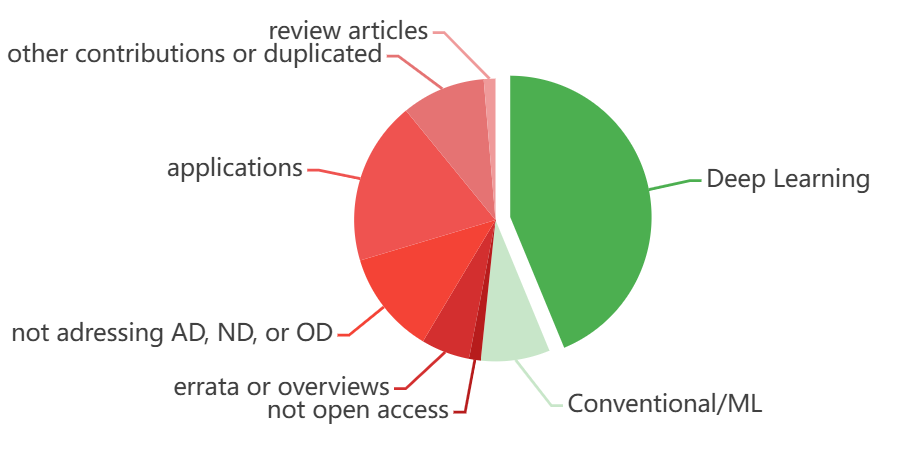}
    \caption{Distribution of the 898 Scopus documents screened from the 2019-2024 period into eligible (green) and excluded (red) categories. Among the 464 eligible, 393 correspond to DL-based methods, the focus of this review. The 434 excluded were based on six criteria (12, 86, 168, 106, 50, 12).}
    \label{fig:piechart}
\end{figure}

Figure~\ref{fig:piechart} shows the distribution of the 898 documents retrieved from the 2019--2024 period after completing the screening and categorization processes. A total of 434 documents were excluded from the analysis based on the above-mentioned criteria. Specifically, 12 were review papers, 70 were other contributions, 16 were duplicates and 168 were application-focused studies. Additionally, 106 documents did not explicitly address the topic of anomaly detection, 50 were errata or overview papers and only 12 were not available through open access or institutional subscriptions. As expected, many of the documents not addressing AD, ND or OD were retrieved through the inclusion of "pattern recognition" term in the search query.
While these studies largely contribute to advancing MTS analysis, via forecasting, clustering or classification strategies, they do not explicitly tackle anomaly, novelty or outlier detection and, therefore, were excluded from this review. In general, title and abstract screening enabled the identification of most documents classified as review papers, those not addressing AD, ND or OD, errata and overview articles, as well as those not available. The remaining documents were assessed in full-text for eligibility.

Among the 464 eligible documents, 393 were categorized as DL-based methods and the remaining 71 were focused on conventional or ML-based approaches. This includes methods relying on architectures such as RNN, Convolutional Neural Network (CNN), Transformers, among others, but also simpler models, such as Multi-layer Perceptron (MLP) models with multiple stacked layers \citep{FlightAD, TRIDENT}. Note that models combining conventional or ML approaches with DL-based optimization procedures were not included in this review, as they were not considered DL-based models for anomaly detection. This included hybrid approaches e.g. combining Self-Organizing Maps with Deep Reinforcement Learning~\citep{Su2023}.

An additional set of 16 papers proposing DL-based methods were identified through manual screening of the reference lists of review articles, resulting in a final set of 409 publications considered in this review. 
These additional papers were retrieved from preprint platforms such as arXiv or published in reputable venues, such as the Proceedings of the AAAI Conference on Artificial Intelligence, published by the Association for the Advancement of Artificial Intelligence (\href{https://aaai.org/}{https://aaai.org/}), only indexed in Scopus from 2023 onwards. Despite their exclusion from Scopus indexation service, these papers exhibit high citation counts and are frequently used as benchmarks for evaluating new MTSAD methods, including TimesNet \citep{TimesNet}, TS2Vec \citep{TS2Vec}, Anomaly Transformer \citep{AnomalyTransformer}, BeatGAN \citep{BeatGAN}, GANF \citep{GANF}, DCdetector \citep{DCdetector} and THOC \citep{THOC} as examples.

\subsection{The proposed taxonomy on MTSAD}

Table~\ref{tab:taxonomy} presents the proposed taxonomy to organize the DL-based approaches aiming at MTSAD. The taxonomy includes eleven different dimensions grouped into three main parts: Input, Output and Model. Briefly, the Input part characterizes how MTS data is fed into the model, while the Output defines how anomalies are represented and identified once the model processes the data. Finally, the Model part aims to capture different aspects of the DL architecture and training setup. 

The \textbf{Input} part of the taxonomy is described using three dimensions: Preprocessing, Format and Content. The Preprocessing (Yes~\textbar~No) dimension indicates whether the raw MTS data were transformed, e.g. to enhance trends, seasonality, shape, or to convert it into another representation, like an image or frequency domain. Standard preprocessing steps, such as data cleaning, normalization or imputation, are not considered here. 
The Format (Point~\textbar~Window~\textbar~Image~\textbar~Window + Image) dimension defines the structure of the input, where \textit{Point} represents the multivariate information at a single time step $t$ in a vector format (i.e., eq.~\eqref{eq:window} with $w{=}0$). \textit{Window} refers to a matrix of multivariate observations in $w{>}0$ consecutive time steps. An \textit{Image} may be seen as a transformed version of a window, often used to emphasize specific patterns. While also matrix-based, its dimensions may differ from the original window, depending on the transformation that is applied to the original data. The two matrix representations may also be combined for into \textit{Window + Image} format.
Finally, the Content (Time~\textbar~Others~\textbar~Time + Others) dimension describes the nature of the information encoded in the input, whether to capture temporal ordering or encode exclusively other information, such as relationships across variables, across time steps or represent frequency-domain characteristics. Temporal ordering may also be combined with other complementary information.

The \textbf{Output} part of the taxonomy includes the dimensions Type, Optimal Threshold, Time-Varying Threshold and Granularity. The Type (Prediction~\textbar~Reconstruction~\textbar~Representation~\textbar~Hybrid) dimension defines how anomaly scores are computed. In the \textit{Prediction}, \textit{Reconstruction} and \textit{Representation} categories, anomaly scores are computed based on prediction error, reconstruction error or deviations in learned representations (either in latent or label space), respectively. The \textit{Hybrid} category combines two or more of the previous strategies to take advantage of their complementary strengths. The dimensions Optimal Threshold (Yes~\textbar~No~\textbar~Not applicable) and Time-Varying Threshold (Yes~\textbar~No~\textbar~Not applicable) are self-explanatory, indicating whether thresholding is data-driven or predefined, manually set or based on heuristics, and whether it adapts over time. The category \textit{Not applicable} refers to methodologies that do not define a Threshold, e.g. those that rely exclusively on non-binary metrics for performance evaluation (see Section \ref{sec: performance_evaluation}). Finally, the Granularity (Point~\textbar~Subsequence) dimension indicates the resolution of anomaly labeling. In the \textit{Point} setting, each time step receives an individual anomaly score or label, whereas in the \textit{Subsequence} setting, labels are assigned to a window or segment of the time series.

\begin{figure}[h!]
    \centering
    \includegraphics[width=0.9\linewidth]{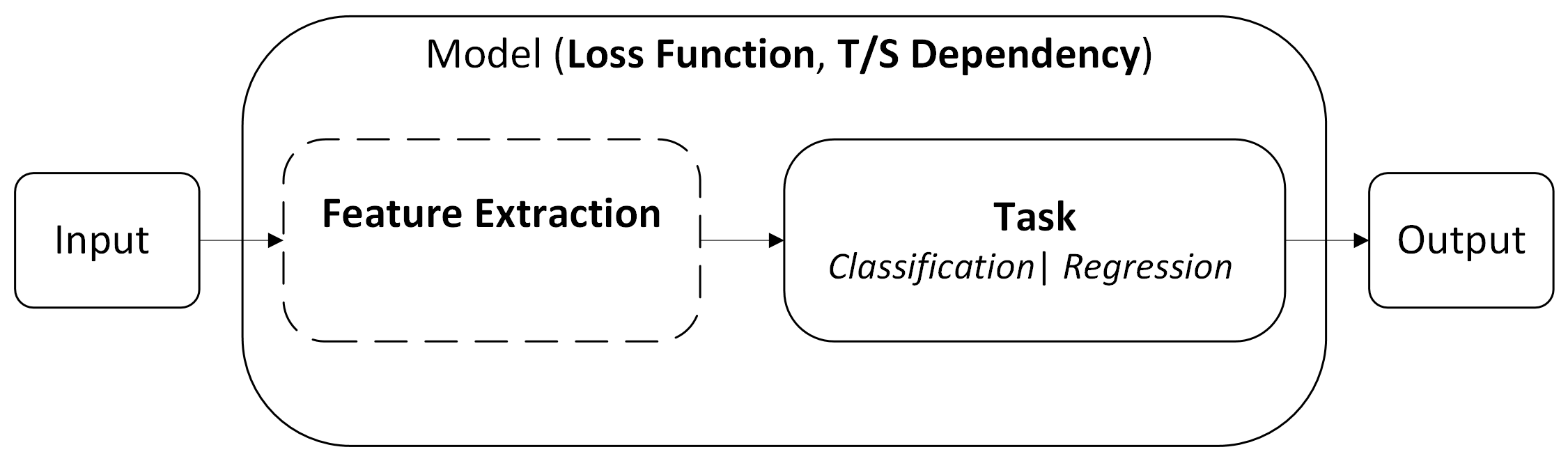}
    \caption{
    Schematic diagram of the Model dimensions (in bold). The dashed border indicates that the Feature Extraction module is optional.}
    \label{fig: feat_extraction}
\end{figure}

The \textbf{Model} part of the taxonomy is described using the dimensions Task, Loss Function, T/S Dependency and Feature Extraction. As illustrated in Figure \ref{fig: feat_extraction}, the model may optionally employ a dedicated Feature Extraction module, whose features are used to perform the Task (classification or regression), with the entire process optimized through a Loss Function under a given T/S Dependency configuration. The Task (Classification~\textbar~Regression) dimension reflects the learning objective, distinguishing models trained for classifying inputs from those predicting continuous values. The Loss Function (Cross-Entropy~\textbar~SumE~\textbar~NLL~\textbar~ELBO~\textbar~Adversarial~\textbar~~Contrastive~\textbar~Multi-objective) dimension captures the optimization criterion used during training. 
The T/S Dependency (Temporal~\textbar~Spatial~\textbar~ Sequential T$\rightarrow$S~\textbar~Sequential S$\rightarrow$T~\textbar~Nested~\textbar~Parallel~\textbar~Others) dimension refers to how the model captures \textit{Temporal} (T) and \textit{Spatial} (S) dependencies. Some models focus exclusively on one dimension, while others process both sequentially (\textit{T$\rightarrow$S} or \textit{S$\rightarrow$T}), jointly in a \textit{Nested}, simultaneously in a \textit{Parallel} fashion, or in alternative structures grouped as \textit{Others}. The Feature Extraction (No~\textbar~Temporal~\textbar~Spatial~\textbar~Spatio-Temporal) dimension indicates whether the model includes an internal module for extracting \textit{Temporal}, \textit{Spatial}, \textit{Spatio-Temporal} features. Models with no such module are labeled as \textit{No}. It is worth noting that the choice of a category in the Feature Extraction dimension does not imply a corresponding category in the T/S Dependency dimension; this distinction will be further elaborated in a subsequent section.

\begin{table}[t!]
\centering
\caption{Proposed taxonomy of DL-based approaches for MTSAD, structured into three main parts and several dimensions that reflect key modeling choices.}
\label{tab:taxonomy}
\footnotesize
\begin{tabular}{c c l p{6cm}}
\toprule
\textbf{Part} & \textbf{Dimension} & \textbf{Categories} & \textbf{Description} \\
\midrule

\multirow{8}{*}{\textbf{Input}}
& Preprocessing & Yes~\textbar~No & Raw TS~\textbar~Transformed TS \\ 
\cmidrule(l){2-4}

& \multirow{4}{*}{Format}
& Point & Single time step $t$ (vector format) \\ 
\cmidrule(l){3-4}
& & Window & Consecutive time steps (matrix format) \\
\cmidrule(l){3-4}
& & Image & Transformed version of a window (matrix format) \\ 
\cmidrule(l){3-4}
& & Window + Image & Combination of Window and Image \\ 
\cmidrule(l){2-4}

& \multirow{3}{*}{Content}
& Time & Temporal ordering \\ 
\cmidrule(l){3-4}
& & Others & Other types of information \\ 
\cmidrule(l){3-4}
& & Time + Others & Temporal ordering combined with other types of information \\
\midrule

\multirow{7}{*}{\textbf{Output}}
& \multirow{4}{*}{Type}
& Reconstruction & Reconstruction error \\ 
\cmidrule(l){3-4}
& & Prediction & Prediction error \\ 
\cmidrule(l){3-4}
& & Representation & Deviations in learned representations \\ 
\cmidrule(l){3-4}
& & Hybrid & Combination of two or more types \\ 
\cmidrule(l){2-4}

& \begin{tabular}[c]{@{}c@{}}Optimal\\ Threshold\end{tabular} & Yes~\textbar~No~\textbar~Not Applicable & Data-driven thresholding strategy \\ 
\cmidrule(l){2-4}
& \begin{tabular}[c]{@{}c@{}}Time-varying\\ Threshold\end{tabular} & Yes~\textbar~No~\textbar~Not Applicable & Threshold adapts over time \\ 
\cmidrule(l){2-4}
& Granularity & Point~\textbar~Subsequence & Resolution of anomaly labeling \\
\midrule

\multirow{20}{*}{\textbf{Model}}
& Task & Classification~\textbar~Regression & Learning objective \\ 
\cmidrule(l){2-4}

& \begin{tabular}[c]{@{}c@{}}Loss\\ Function\end{tabular}
& \begin{tabular}[c]{@{}l@{}l@{}}
Cross-Entropy~\textbar~SumE\\ NLL~\textbar~ELBO~\textbar~Adversarial \\Contrastive~\textbar~Multi-objective
\end{tabular}
& Optimization criterion \\ 
\cmidrule(l){2-4}

& \multirow{12}{*}{\begin{tabular}[c]{@{}c@{}}T/S\\ Dependency\end{tabular}}
& Temporal & Only handles temporal dependencies \\ 
\cmidrule(l){3-4}
& & Spatial & Only handles spatial dependencies \\ 
\cmidrule(l){3-4}
& & Sequential T$\rightarrow$S & Handles temporal dependencies first, followed by spatial dependencies \\ 
\cmidrule(l){3-4}
& & Sequential S$\rightarrow$T & Handles spatial dependencies first, followed by temporal dependencies \\ 
\cmidrule(l){3-4}
& & Nested & Handles both dependencies, embedding one within the other \\ 
\cmidrule(l){3-4}
& & Parallel & Handles both dependencies simultaneously, in a parallel fashion  \\ 
\cmidrule(l){3-4}
& & Others & Handles both dependencies but not sequentially, in a parallel nor in a nested structure\\ 
\cmidrule(l){2-4}

& \multirow{4}{*}{\begin{tabular}[c]{@{}c@{}}Feature\\ Extraction\end{tabular}}
& No & No feature extraction module \\ 
\cmidrule(l){3-4}
& & Temporal & Temporal feature extraction module\\ 
\cmidrule(l){3-4}
& & Spatial & Spatial feature extraction module\\ 
\cmidrule(l){3-4}
& & Spatio-Temporal & Spatio-temporal feature extraction module \\
\bottomrule
\end{tabular}
\end{table}

\subsection{Insights from prior research}

The dimensions of the taxonomy were derived from the analysis of the screened documents and complemented by insights from review papers \citep{Choi2021,Correia2024,Li2023,ZamanzadehDarban2024}. 
This approach allowed to derive a unified taxonomy upon prior work while refining it to reflect the current state-of-the-art. Table \ref{tab:taxonomy_comparison} presents literature contributions mapped into the dimensions of the proposed taxonomy.
Regarding the \textbf{Input} part, \citet{ZamanzadehDarban2024} proposed a categorization focused on the \textit{Format} dimension, distinguishing between \textit{Point} and \textit{Window}; however, screening methodological papers revealed these categories are insufficient to fully characterize MTSAD inputs, motivating the addition of an \textit{Image} category to \textit{Format} and the introduction of the new dimensions \textit{Preprocessing} and \textit{Content}. 
The dimensions of the \textbf{Output} part were also defined based on insights from prior reviews \citep{Choi2021,Li2023,ZamanzadehDarban2024}. \citet{Choi2021} groups the methodologies based on how the anomaly score is computed (either via reconstruction error, prediction error or dissimilarity) while \citet{ZamanzadehDarban2024} categorized as forecasting-based, reconstruction-based, representation-based or hybrid. The dissimilarity category broadly aligns with the representation-based category, as both rely on comparisons in transformed feature spaces. Therefore, the categories from \citet{ZamanzadehDarban2024} were adopted for the \textit{Type} dimension. For the \textit{Threshold} dimension, \citet{Li2023} proposed an extensive taxonomy including \textit{fixed}, \textit{optimal}, \textit{adaptive}, \textit{specific} and \textit{non-parametric} thresholds. However, this scheme mixes characteristics with computation methods and includes overlapping categories. The current taxonomy simplifies it by proposing two new binary dimensions, \textit{Optimal Threshold} and \textit{Time-varying Threshold}. Lastly, the \textit{Granularity} dimension captures the level at which the anomaly detection output is produced, distinguishing between individual time points and (possibly overlapping) sequences of time steps. This differs from the categorization of anomaly types, where (point, subsequence) \citep{ZamanzadehDarban2024} or (point, interval, series) \citep{Li2023} describe the anomaly type rather than the output resolution.

Existing surveys often focus on categorizing the MTSAD models solely based on their underlying architecture \citep{Choi2021, ZamanzadehDarban2024}, yet several important modeling aspects should be considered. This review proposes an extended view of the \textbf{Model} part, by incorporating the new \textit{Loss Function} dimension and by expanding the categories in  \textit{Feature Extraction} and \textit{T/S dependency}. \citet{Choi2021} discuss various types of loss, including \textit{adversarial}, \textit{reconstruction}, \textit{prediction} and \textit{negative log-likelihood} losses. In the new taxonomy, the \textit{Loss Function} dimension captures the mathematical form of the loss, while the \textit{Type} dimension identifies the components used to compute the dissimilarity.
Regarding \textit{Feature Extraction}, the new taxonomy extends previous work \citep{Li2023} by not only indicating whether feature extraction is used but also detailing the specific types of features extracted. Finally, the \textit{T/S Dependency} dimension expands the existing categorizations \textit{Temporal}, \textit{Spatial} and \textit{Spatio-temporal} \citep{ZamanzadehDarban2024} by providing a more detailed and structured view of how spatio-temporal dependencies are modeled and the emphasis each method places on them. This includes approaches based on \textit{Sequential}, \textit{Nested}, \textit{Parallel} or \textit{Other} paradigms. Finally, although \citet{Correia2024} do not explicitly define a taxonomy, they address related aspects that align with the new one: \textit{sliding windows} and \textit{input sequence} align with the \textit{Input Format} dimension; \textit{Prediction} and \textit{Reconstruction} relate to \textit{Output Type}; and model categories (AE, RNN, CNN, Transformers) connect indirectly to \textit{Model Feature Extraction} or \textit{Model T/S dependencies}.

\begin{table*}[ht]
\centering
\caption{Literature contributions to the novel of taxonomy proposed in this work.
--- indicates it is absent or not explicitly defined. }
\label{tab:taxonomy_comparison}
\footnotesize
\begin{tabular}{l c p{2cm} p{2cm} p{2cm} p{2cm}}
\toprule
\textbf{Part} & \textbf{Dimension} & \textbf{\citet{Choi2021}} & \textbf{\citet{Li2023}} & \textbf{\citet{ZamanzadehDarban2024}} & \textbf{\citet{Correia2024}}  \\
\midrule
\multirow{3}{*}{\textbf{Input}}
& Preprocessing  & ---  & --- & --- & --- \\
\cmidrule(l){2-6}
& Format & ---  & ---  & Point, Window  & Sliding window, Input sequence  \\
\cmidrule(l){2-6}
& Content  & --- & --- & --- & --- \\
\midrule
\multirow{5}{*}{\textbf{Output}}
& Type & Reconstruction, Prediction, Dissimilarity & --- & Forecasting, Reconstruction, Representation, Hybrid & Prediction, Reconstruction \\
\cmidrule(l){2-6}
& \begin{tabular}[c]{@{}c@{}}Optimal\\ Threshold\end{tabular}
  & \multirow{2}{2cm}{---}
  & \multirow{2}{2cm}{Fixed, Optimal, Adaptive, Specific, Non-parametric}
  & \multirow{2}{2cm}{---}
  & \multirow{2}{2cm}{---} \\
  \cmidrule(l){2-2}
& \begin{tabular}[c]{@{}c@{}}Time-variyng\\ Threshold\end{tabular} & & & & \\
\cmidrule(l){2-6}
& Granularity  & --- & Point, Interval, Series & Point, Subsequence & --- \\
\midrule
\multirow{4}{*}{\textbf{Model}}
& Task & --- & --- & ---& ---  \\
\cmidrule(l){2-6}
& \begin{tabular}[c]{@{}c@{}}Loss\\ Function\end{tabular}  & Adversarial, Reconstruction, Prediction, NLL & --- & ---& ---  \\
\cmidrule(l){2-6}
& \begin{tabular}[c]{@{}c@{}}T/S\\ Dependency\end{tabular}& --- & ---  & Temporal, Spatial, Spatio-temporal   & --- \\
\cmidrule(l){2-6}
& \begin{tabular}[c]{@{}c@{}}Feature\\ Extraction\end{tabular} & --- & Binary (yes/no) & --- & --- \\
\bottomrule
\end{tabular}
\end{table*}

\section{Overview of the Literature on the Dimensions of the New Taxonomy}\label{sec4}

\begin{figure}[h]
    \centering
    \includegraphics[width=\linewidth]{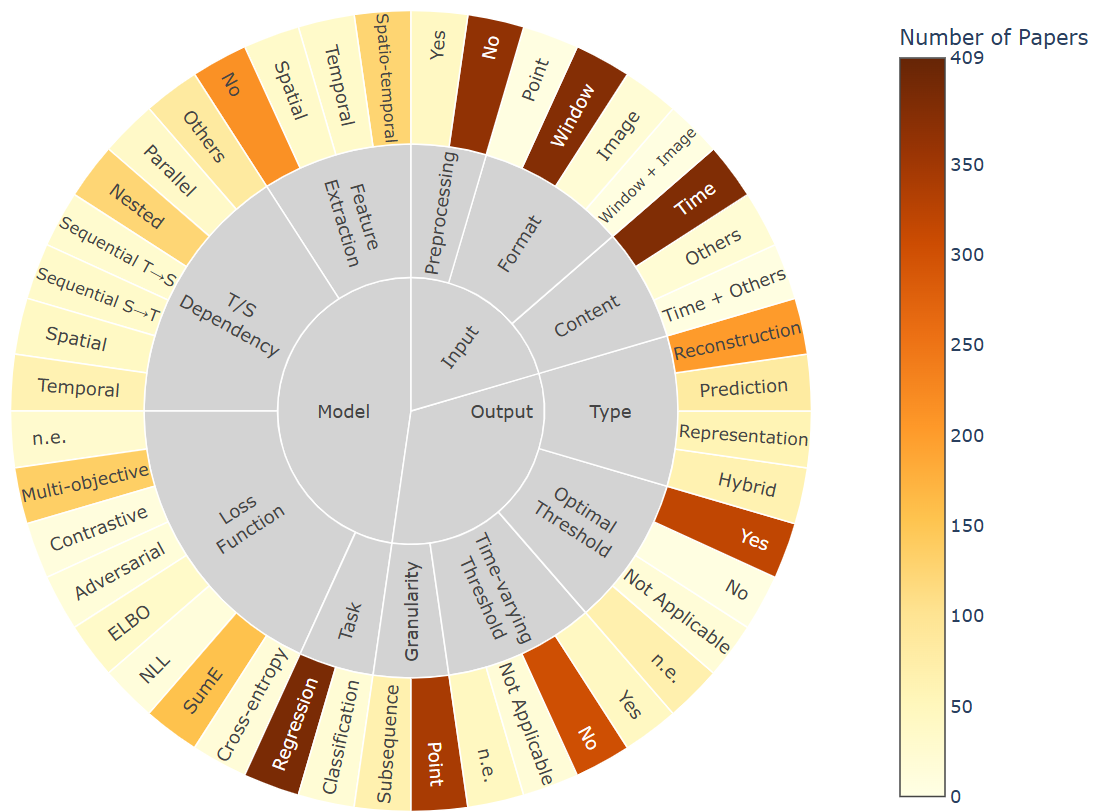}
    \vspace{0.05cm}
    \caption{Sunburst chart of 409 categorized documents across the taxonomy. The inner circle shows the Parts, followed by Dimensions and Categories. Colors represent the number of documents and ``n.e.'' (not explicit) refers to papers that did not explicitly report information for a given dimension. The interactive version of this figure is available at \url{https://ieeta-pt.github.io/DL-MTSAD/Sunburst.html}.}
    \label{fig:sunburst}
\end{figure}

Figure~\ref{fig:sunburst} presents the classification of the 409 categorized models according to the proposed taxonomy. The three Parts of the taxonomy are shown at the innermost circle of the Sunburst. Moving outward, the eleven dimensions of the taxonomy are displayed and the outermost layer corresponds to the specific categories within each dimension. Each model is assigned to exactly one category per dimension. Colors in the chart indicate the number of documents assigned to each category, providing a visual sense of their relative prevalence. For some dimensions, such as Loss Function, Time-varying Threshold and Optimal Threshold, certain papers did not provide explicit information; these cases are represented by the category labeled ``n.e.'' (not explicit). An interactive version of this figure is available at \url{https://ieeta-pt.github.io/DL-MTSAD/Sunburst.html}, allowing readers to consult the counts per category and to explore the specific parts and dimensions of the taxonomy. The following sections provide a more detailed overview of the literature aligned with each dimension of the taxonomy, with representative examples of each category cited throughout. The complete list of categorized papers is also available at \url{https://github.com/ieeta-pt/DL-MTSAD}.

\subsection{Input: Preprocessing, Format and Content}

The dimensions of the Input include Preprocessing, Format and Content. Regarding \textbf{Content} (Time~\textbar~Others~\textbar~Time + Others), it is clear that most studies preserve and make use of the temporal ordering of the time series, reflecting its fundamental importance for anomaly detection. Temporal data is typically used independently, and just a subset of studies enhances it with additional attributes such as spatial correlations, temporal dependencies or frequency-driven features. Some studies focus exclusively on non-temporal information, with spatial correlations being the most common, highlighting the role of relationships among variables in multivariate anomaly detection. 
Temporal-only inputs are simpler and computationally efficient, but may fail to capture inter-metric anomalies, whereas incorporating spatial or hybrid content can improve detection performance at the cost of
increased model complexity and potential noise, when the additional content is weakly informative.
In terms of \textbf{Format} (Point~\textbar~Window~\textbar~Image~\textbar~Window + Image), the majority of inputs are structured as \textit{Window}, which preserve temporal context and are necessary for the detection of Collective and Contextual anomalies, whereas only a few studies make use of time \textit{Point}, which are suitable to detect Point anomalies. Although most studies utilize a sliding window technique to extract fixed-length windows, temporal segmentation can alternatively be applied, resulting in non-overlapping segments with different lengths (e.g. \citet{seq2GMM}). Moreover, some studies convert MTS windows into \textit{Image} representations to capture temporal or spatial correlations, and just a few combine both \textit{Window} and \textit{Image} formats to enrich the input with complementary information. Image representations enable the use of models typically used for image analysis, such as CNN, but at the cost of increased preprocessing complexity and potential loss of temporal resolution.

With respect to \textbf{Preprocessing} (Yes~\textbar~No), most studies do not apply any preprocessing step and instead rely directly on raw time series as input. When preprocessing is applied, it typically emphasizes temporal characteristics of the data through filtering, decomposition or multi-scale transformations \citep{SDDformer, Meta-MWDG, TFAD}. Other strategies focus on frequency-based representations or on explicitly computing spatial and temporal correlations, either alone or combined with raw series, to enrich the input of the DL-model \citep{STFT-TCAN}. While preprocessing can improve detection by highlighting relevant signal characteristics, it also introduces additional design choices and may hide anomalies that exist in the raw time series.

The three Input dimensions were defined independently, and the analysis of the categorized papers shows that most MTSAD methods rely on raw MTS in \textit{Window} format, maintaining the inherent temporal ordering. However, the categorization shows other clear associations between these dimensions, as changes in \textbf{Content} or \textbf{Format} usually require \textbf{Preprocessing}, and \textbf{Format} alterations are often tied to \textbf{Content} changes.
\textbf{Preprocessing} techniques are closely related to the \textbf{Content} of MTS, as they can be used to emphasize or extract specific aspects of the data that are relevant for AD. Preprocessing techniques that preserve the temporal content of MTS often aim to highlight key aspects such as trend, seasonality or multi-scale structures. Examples include extracting trend and residual components using moving averages, as in SDDformer (Separation, Decomposition, and Dual Transformer-based autoencoder) \citep{SDDformer}; seasonal-trend decomposition via STL, as in STL-ConvTransformer \citep{STL-ConvTransformer}; Hodrick–Prescott filtering, as in TFAD (Time-Frequency analysis based time series Anomaly Detection) \citep{TFAD}; and multi-scale decomposition via Discrete Wavelet Transforms (DWT), as in Meta-MWDG \citep{Meta-MWDG}. Other preprocessing techniques focus on frequency-based content, transforming time-domain sequences into the frequency-domain to capture amplitude/power information relevant for detecting anomalies, particularly seasonal patterns. For instance, the STFT-TCAN model employs Short-Time Fourier Transforms (STFT) to extract amplitude features and integrates them with the original temporal data to enrich the representation of the model input \citep{STFT-TCAN}.

\textbf{Preprocessing} techniques are often used to modify the input \textbf{Format}, e.g. transforming MTS from \textit{Window} into \textit{Image} representations to better capture temporal or spatial patterns. Traditional approaches for imaging UTS include Recurrence Plots (RP), Markov Transition Fields (MTF) and Gramian Angular Fields (GAF) \citep{RP, GAF_and_MTF}. Since these methods are univariate, they are typically applied to each metric individually to extract temporal correlations. For example, T2IAE (Time-series to Image-transformed adversarial Autoencoder) converts a window sequence with $m$ metrics into $m$ images, which are then stacked and fed to the DL-model \citep{Kang2024}. Moreover, some MTSAD models use GAF to capture spatial correlations among metrics by computing angular differences, producing an $M \times M$ matrix for each time point \citep{GAF-GAN, MDSCAD}. Other studies employ signature matrices to transform \textit{Window} into \textit{Image}, encoding spatial correlations of a subsequence of size $w$ as  
\begin{equation}
r_{i,j} = \frac{\sum_{\delta=0}^{w} x_{i,t-\delta} x_{j,t-\delta}}{\kappa}, \quad i,j=1,2,...,M \ ,
\end{equation}
where $\kappa$ is typically set to $w$ but may vary depending on the similarity measure used, such as correlation or cosine similarity \citep{MSCRED, DCGAN, MCRAAD, FMUAD}. Additionally, some approaches generate sets of snapshot images from sliding windows over the MTS, providing a sequence of input images for the DL-model \citep{AC-LSTM, CANet, MAC-Net}.

Different \textbf{Preprocessing} techniques can also be used to combine multiple \textbf{Formats} and \textbf{Content} types in the Input. For instance, FMUAD (Forecast-based, Multi-aspect, Unsupervised AD) computes signature matrices using the cosine similarity between each pair of metrics to capture spatial correlations, and applies the Discrete Fourier Transform (DFT) to extract frequency amplitudes from each metric. These two image-based representations, along with the original temporal data, are then processed separately through dedicated modules to model correlation changes, temporal pattern variations and spatial dynamics \citep{FMUAD}.

\subsection{Output: Type, Thresholding and Granularity}

Important aspects of the Output include Type, Thresholding and Granularity. The Output \textbf{Type} (Prediction~\textbar~Reconstruction~\textbar~Representation~\textbar~Hybrid) defines how the anomaly score $AS$ is computed. The most common is \textit{Reconstruction}, followed by \textit{Prediction}, with \textit{Hybrid} and \textit{Representation}-based methods being less frequent. \textit{Reconstruction} methods, such as OmniAnomaly and Interfusion, compute $AS$ from reconstruction errors \citep{OmniAnomaly, Interfusion}, USAD (UnSupervised AD on multivariate time series) combines the $AS$ of two adversarial autoencoders \citep{USAD}, and MSCRED (Multi-Scale Convolutional Recurrent Encoder-Decoder) uses poorly reconstructed elements in signature matrices \citep{MSCRED}. Prediction methods, like GDN (Graph Deviation Network) and GTA (Graph learning with Transformer framework for Anomaly detection), compute errors between predicted and actual values \citep{GDN, GTA}, while Representation approaches, including TFAD and MTGFlow (Multivariate Time series anomaly detection via dynamic Graph and entity-aware normalizing Flow), compare learned representations or likelihoods \citep{TFAD, MTGFlow}. Each type has strengths: briefly \textit{Reconstruction} captures overall patterns, \textit{Prediction} detects sudden changes, and \textit{Representation} captures latent deviations but can be sensitive to noise \citep{MTAD-GAT, CAE-M, USAD}. \textit{Hybrid} methods combine types to mitigate weaknesses, e.g. MTAD-GAT (Multivariate Time-series Anomaly Detection via GAT) sums reconstruction and prediction errors, and MEMTO (Memory-guided Transformer for MTSAD) multiplies latent and input space deviations \citep{MTAD-GAT, MEMTO}.

Another crucial aspect in the MTSAD model is how the threshold is defined (see \eqref{eq: threshold}) as it directly affects the performance of the model. An \textbf{Optimal Threshold} (Yes~\textbar~No) is data-driven, which occurs in almost all the categorized documents. A \textbf{Time-varying Threshold} (Yes~\textbar~No) adapts over time according to contextual characteristics of the data. Most of the screened papers use a constant threshold rather than a time-varying one. A subset of papers choose not to define a threshold, instead employing evaluation metrics that are independent of thresholds (see Section \ref{sec: performance_evaluation}) \citep{GATAMAF, GST-Pro}. Some papers do not explicitly mention whether the implemented thresholds are Time-varying or Optimal. 

These two dimensions describe different aspects of the threshold. A non-optimal threshold is typically constant, often predetermined based on insights from experts. An \textbf{Optimal Threshold} can be either Time-varying or not. For instance, a time-invariant threshold can be computed based on the training or validation sets by either getting the maximum $AS$ obtained on that dataset or by using the sigma-rule  

\begin{equation}\label{eq: sigma_rule}
    \delta = \mu + z\sigma,
\end{equation}

\noindent where \(\mu\) and \(\sigma\) denote the mean and standard deviation of the $AS$, respectively. The parameter \(z\) is a tunable hyperparameter that controls the sensitivity by determining how far a score must deviate from the mean to be considered anomalous. Another typical approach is to designate a specific portion of the training (or validation) set as anomalies and determine $\delta$ as the threshold that best distinguishes these anomalies based on their $AS$.

Among the most commonly adopted techniques for threshold selection is the Peaks-Over-Threshold (POT) approach, derived from the second theorem of Extreme Value Theory (EVT). 
According to this theorem, the distribution of data points that exceed a sufficiently high threshold can be approximated by a Generalized Pareto Distribution (GPD) \citep{OmniAnomaly,POT_and_variants}. The GPD distribution function can be written as

\begin{equation}
{F}(s) = P (S-th \leq s|S>th) \sim 1-\left( 1+\frac{\gamma s}{\beta} \right)^{-\frac{1}{\gamma}}
\end{equation}

\noindent where \(th\) is a given threshold, S is a random variable and \(\gamma\) and \(\beta\) are shape and scale parameters of GPD, typically estimated using the Maximum Likelihood Estimate (MLE). The refined anomaly threshold, $\delta$ can be computed by

\begin{equation}
\delta \simeq th + \frac{\hat{\beta}}{\hat{\gamma}} \left( \left( \frac{qN}{N_{th}} \right)^{-\hat{\gamma}}-1 \right),
\end{equation}

\noindent where $q$ is the desired proportion of anomalies, $N$ is the number of AS values, $N_{th}$ is the number of these values that are higher than $th$ and $\hat{\gamma}$ and $\hat{\beta}$ are estimates for model parameters \citep{OmniAnomaly,POT_and_variants}. Therefore, using this method, two parameters need to be tuned: $q$ and the threshold $th$. Alternatively, some works use ML-based algorithms for anomaly/outlier detection on top of the anomaly scores or features computed by the DL models. Some examples include the IF, OCSVM and the Local Outlier Factor (LOF) \citep{Nguyen2021,CSL}.
Finally, several works perform an exhaustive search over a predefined range of potential thresholds to identify that with the highest F1-score. In this approach, the F1-score is calculated for each candidate threshold in the range, and the optimal threshold \(\delta\) is selected as the one maximizing the F1-score \citep{USAD, GTA, Interfusion, Zhou2022_CAE-AD}.
This approach provides an upper limit to the performance of the method. However, contrary to POT and ML-based approaches, it is not feasible in an online setting, where the threshold needs to be set before inference, without access to labeled anomalies.

Instead of fixing a threshold for the entire inference stage, other approaches have been adapting a \textbf{Time-varying Threshold}. For example, the sigma-rule \eqref{eq: sigma_rule} can be used in a time-varying fashion with an Exponential Weighted Moving Average (EWMA) and an Exponential Weighted Standard Deviation (EWSD) \citep{Matar2023}, or with the Nonparametric Dynamic Thresholding (NDT) method proposed by \citet{MSL}. Finally, variations of POT may also be used to adapt the threshold over time, including streaming POT (SPOT) and drift-aware SPOT (DSPOT) \citep{POT_and_variants}. Compared to a time-invariant threshold, a time-varying threshold has the advantage of dynamically adjusting to the data, which is useful when the data is non-stationary. However, it has the disadvantage of needing to be computed continuously and may be sensitive to noise, as fluctuations in the \textit{AS} may adapt the threshold unnecessarily, reducing detection performance.

Most of the Output \textbf{Granularity} (Point~\textbar~Subsequence) choices are \textit{Point} rather than \textit{Subsequence}. Even when $AS$ is computed at the window level, it is usually mapped to points — commonly the last \citep{OmniAnomaly, USAD, CTAD}, but sometimes the first \citep{ZhouX2023LAT} or middle point \citep{WangF2023}. Point-level $AS$ can also be estimated by averaging overlapping windows \citep{MAD-GAN, TFAD}. In contrast, Subsequence methods label entire windows directly \citep{MSCRED, CAE-M, AMSL}. 
While subsequence granularity treats the entire window as the detection
unit, point granularity enables finer temporal resolution, allowing more
precise localization of anomalies.
Although granularity is independent of evaluation, point-level approaches often use point-adjustment to align predictions with segment-level ground truth. 
Moreover, beyond detecting anomalies at the point or subsequence level, some studies also identify the specific metrics where anomalies occur, enabling fine-grained anomaly detection \citep{MFGAD, STVAE}.

\subsection{Model: Task and Loss Function}

The model \textbf{Task} (Regression~\textbar~Classification) for anomaly detection is usually formulated as \textit{Regression} rather than \textit{Classification}. Anomalies are rare by nature, leading to highly imbalanced datasets, which makes supervised classification challenging because it requires sufficient labeled examples of both normal and anomalous events. Moreover, regression settings do not need labeled data on anomaly for training. Instead, they are typically trained on datasets assumed to be predominantly normal, relying on the fact that anomalies are rare \citep{USAD, OmniAnomaly}.
Additionally, while classification models directly output a label indicating the presence of an anomaly, regression models compute an $AS$, providing a measure of anomaly severity. However, classification models avoid the need to define additional parameters, such as thresholds, and in the presence of high-quality labeled data, can perform multi-class anomaly detection \citep{PCAC, MBead, Qi2023}, distinguishing between different types of anomalies, something regression models cannot achieve.

Another key aspect of model training is the optimization criterion, where the \textbf{Loss Function} is used to compute the average loss (or cost) over the entire dataset as presented in \eqref{eq:cost_function}. The most frequently used loss functions belong to the \textit{SumE} category, while \textit{Multi-objective} formulations have also gained significant popularity. Less common alternatives include \textit{Cross-Entropy}, \textit{NLL}, \textit{ELBO}, \textit{Adversarial} and \textit{Contrastive} losses. Finally, a subset of papers does not explicitly report the mathematical form of the loss function considerd in training. 

\textit{Cross-entropy} loss functions are among the most widely used in classification tasks, particularly when labels are represented in a one-hot form (i.e. when each class label is represented as a vector of length $C$ with all entries equal to 0 except for a single 1 at the index $c=\{1,2,\cdots,C\}$ corresponding to the correct class). 
The general formulation is the Categorical Cross-Entropy (CCE) that is commonly applied in multi-class scenarios
\begin{equation}
\mathcal{L}_{\text{CCE}} = - \sum_{c=1}^{C} \mathbf{\mbox{1}}_c(y) \log(\hat{p}_c),
\end{equation}
where $C$ is the number of classes, $\mathbf{\mbox{1}}_c(y)$ is the indicator function that returns one if $y = c$ and zero otherwise, and $\hat{p}_c$ is the predicted probability for the example $y$ to belong to class $c$. 
The Binary Cross-Entropy (BCE) is the particular case of CCE for $C=2$ and the summation in the CCE reduces to two terms, corresponding to the positive ($c=1$) and negative ($c=2$) classes with probabilities $\hat{p}_1$ and $\hat{p}_2=1-\hat{p}_1$, respectively. 

On the other hand, the loss functions adopted in regression settings are typically based on a transformation $f(\cdot)$ of the difference between the observed data $\textbf{X}$ and model output $\hat{\textbf{X}}$, 
\begin{equation}
\mathcal{L}_{SumE} =  f(\mathbf{X} -\hat{\mathbf{X}}).
\end{equation}
These functions are grouped under the category \textit{SumE} and common choices for $f(\cdot)$ include distance-based functions such as the Mean Absolute Error (MAE) defined as

\begin{equation}
\mathcal{L}_{\text{MAE}} = \frac{1}{Mw} \left\| \mathbf{X} -\hat{\mathbf{X}} \right\|_{(1,1)},
\end{equation}
where $\mathbf{X}$ and $\hat{\mathbf{X}}$ are matrices with $M$ metrics and $w$ time points. The Mean Squared Error (MSE) is obtained equivalently by replacing $\left\| \cdot \right\|_{(1,1)}$ with the Frobenius norm $\left\| \cdot \right\|_F^2$ and the Root MSE (RMSE) corresponds to its square root. In the vector case ($w=1$), the $\left\| . \right\|_{(1,1)}$ is replaced by the Manhattan distance $||\cdot||_1$, while the Frobenius norm is replaced by the Euclidean norm $||\cdot||_2$. Among these options, MSE is usually selected because it is mathematically convenient, as it is differentiable, which makes optimization using gradient-based methods efficient and stable \citep{Terven2025}. Finally, the $SumE$ category also includes less common loss functions, such as the quantile loss \citep{Rancoders} and the Huber loss \citep{Elhalwagy2022, Xu2023}.

The \textit{Negative Log-Likelihood (NLL)} is a general loss function that measures how well a predicted probability distribution matches the observed data, being defined as
\begin{equation}
\mathcal{L}_{\mathrm{NLL}} = - \log p_{\theta}(y \mid x),
\end{equation}
where $p_{\theta}(y \mid x)$ is the model-predicted probability function of $y$ given the input $x$ and the model parameters $\theta$. In classification with one-hot encoding, NLL is equivalent to CCE or BCE. In regression, the choice of $p_\theta$ is usually assumed as a Normal (gaussian) distribution. Minimizing the NLL is equivalent tomaximize the likelihood of the training data, encouraging the model to assign high probability to normal samples \citep{Terven2025}. This approach is commonly followed to train Normalizing Flow (NF) models (e.g. MTGFlow \citep{MTGFlow}), which will be described in the next section.

Concepts borrowed from information theory can also be applied as loss functions. One such example is the Kullback-Leibler (KL) divergence, used to measure the divergence between a probability distribution $q$ and a target distribution $p$, as

\begin{equation}
\mathrm{KL}(p \mid q)=\sum_{i=1}^n p\left(x_i\right) \log \left(\frac{p\left(x_i\right)}{q\left(x_i\right)}\right),
\end{equation}
where $x_i$ is the $i$-th example and $n$ is the total number of points or discrete outcomes. 
The KL divergence is also a key component of the \textit{Evidence Lower BOund (ELBO)}, commonly maximized when training a VAE, providing a lower bound on the data log-likelihood \citep{Terven2025}. The ELBO is defined as
\begin{equation}
\mathcal{L}_{\text{ELBO}}(x) = \mathbb{E}_{q(z \mid x)} \big[\log p(x \mid z) \big] - \mathrm{KL}\big(q(z \mid x) \,\|\, p(z)\big),
\end{equation}
where $\mathbb{E}\big[\cdot\big]$ is the expected value. The first term is the reconstruction term, measuring how accurately the model can reconstruct the input $x$ from the latent variable $z$, and the second term penalizes the divergence between the approximate posterior $q(z|x)$ and the prior $p(z)$. 
If $p_{\theta}(y \mid x)$ is assumed to be Gaussian with fixed variance, then the reconstruction term is proportional to the negative MSE and thus, in practice, this term is often implemented as the MSE loss \citep{OmniAnomaly,Fahrmann2022}.

\textit{Adversarial} and \textit{Contrastive} loss functions are commonly used to enhance model training and representation learning. \textit{Adversarial} loss frames training as a minimax game between a Generator ($G$) and a Discriminator ($D$), where $D$ distinguishes real from generated data, and $G$ aims to generate samples indistinguishable from real ones \citep{GAN}. This encourages $G$ to closely match the true data distribution, 
\begin{equation}
 \min_G \max_D V(D, G) = \mathbb{E}_{x \sim p_\text{data}(x)} [\log D(x)] + \mathbb{E}_{z \sim p_z(z)} [\log (1 - D(G(z)))].
\end{equation}
where $D(\cdot)$ is the probability of a sample to be real, $x$ represents a real sample and $G(z)$ is a sample generated from a latent variable $z$.
The first term (maximized by $D$) measures how well $D$ distinguishes real data sampled from the true distribution $p_\text{data}(x)$, and the second term (maximized by $D$ and minimized by $G$) measures how well $D$ can distinguish generated samples drawn from $p_z(z)$ \citep{GAN, Terven2025}.
\textit{Contrastive} losses, on the other hand, guide the model to differentiate between similar and dissimilar examples. Positive pairs are generated via augmented views of the same data, while negative pairs are created from distorted or shuffled segments  \citep{Contrastive}. The loss maximizes similarity for positive pairs and minimizes it for negative pairs, with common examples including InfoNCE \citep{InfoNCE} and NT-Xtent \citep{NT-Xtent}. Alternative measures, such as KL divergence, can also quantify dissimilarity; for example, DCdetector (Dual attention Contrastive representation learning anomaly detector) uses KL divergence between representations from two temporal views \citep{DCdetector}.

Finally, a widely used approach is to combine different loss functions in a \textit{Multi-objective} formulation, using a composition of multiple (complementary) loss functions. For example, GraphAD combines MSE with the mutual information between two representations, a stable one and a volatility one \citep{GraphAD}. Moreover, MTAD-GAT combines the forecasting and reconstruction losses, computed respectively from RMSE and ELBO. Several studies have been combining the MSE with other losses, e.g. contrastive ones. For instance, the Anomaly Transformer model \citep{AnomalyTransformer} integrates the reconstruction loss (MSE) with the association discrepancy, computed as the symmetrized KL divergence between two data representations, a strategy later adopted by several models \citep{SIET, SSAD}. Similarly, CAE-AD (Contrastive Autoencoder for Anomaly Detection) combines reconstruction loss with contextual and instance-level contrastive losses, operating at the window and timestep scales, respectively \citep{Zhou2022_CAE-AD}.

The choice of loss function defines the optimization criterion used to train the model. \textit{SumE} and \textit{Cross-Entropy} losses minimize the error directly, making them simple and computationally efficient, but they do not account for uncertainty. \textit{NLL} and \textit{ELBO} address this by modeling the data distribution explicitly, enabling uncertainty quantification, at the cost of requiring distributional assumptions. Moving further from direct supervision, the \textit{Adversarial} loss implicitly learns normality through a minimax game, allowing the model to capture complex data distributions, but it introduces training instability. \textit{Contrastive} losses take a different perspective, shifting the focus from error minimization to representation, ensuring that normal samples cluster together in latent space. Finally, \textit{Multi-objective} formulations combine complementary losses to capture multiple aspects of normality, at the cost of increased complexity.

\subsection{Model: T/S dependency and Feature Extraction}

Two other important dimensions of the Model part are T/S dependency and Feature Extraction.
Although T/S Dependency and Feature Extraction relate to the same underlying DL architectures, they capture different aspects of model design. T/S Dependency describes how the model globally organizes the modeling of temporal and spatial dependencies, while Feature Extraction indicates wheter such a dedicated module is employed to extract temporal, spatial, or spatio-temporal features.

Building on existing taxonomies for DL-based methods \citep{ZamanzadehDarban2024, Choi2021}, popular architectures for MTSAD were grouped into temporal architectures comprising RNN, Transformer and TCN, and spatial architectures comprising CNN, GNN, AE, VAE and GAN. 
Other generative architectures, such as Normalizing Flows and Diffusion Models, have also been used in MTSAD despite not directly modeling temporal or spatial dependencies. Normalizing Flows (e.g. MAF and Real NVP) transform a simple base distribution (e.g. Gaussian) into a complex target distribution via invertible mappings, allowing efficient density estimation \citep{Kobyzev2021}. In MTSAD, anomalies are assumed to occur in low-density regions \citep{GANF, MTGFlow}. Diffusion Models (e.g. DDPM) instead corrupt inputs with noise and learn to reverse this process for reconstruction \citep{ImDiffusion, DDTAD}. In practice, these generative models are often combined with spatio-temporal architectures to better capture the complex MTS patterns.

The \textbf{Feature Extraction} dimension (No~\textbar~Temporal~\textbar~Spatial~\textbar~Spatio-Temporal) describes whether MTSAD models employ a dedicated module for extracting features. While all models generate deep representations from the MTS, some explicitly target \textit{Temporal}, \textit{Spatial} or \textit{Spatio-Temporal} features through a specialized module, which is typically composed of the aforementioned DL architectures. Others, in contrast, rely on a unified approach that does not differentiate between these aspects (\textit{No}). With respect to this dimension, almost half of the surveyed approaches omit such a stage altogether, opting instead to intertwine feature learning with prediction or reconstruction, as seen in OmniAnomaly \citep{OmniAnomaly}. However, when feature extraction modules are explicitly introduced, they predominantly target spatio-temporal features, underlining that both space and time information are key to performing accurate anomaly detection. For example, GTA integrates dilated and graph convolutions to capture spatio-temporal representations that a Transformer subsequently processes \citep{GTA}, while MTAD-GAT makes use of two parallel GNN-based modules followed by an RNN to extract spatio-temporal features that feed both prediction and reconstruction branches \citep{MTAD-GAT}. By comparison, models that restrict feature extraction to only spatial or temporal aspects are less common and more specialized: GDN extracts spatial features through a GNN-based module \citep{GDN}, CAE-M (Convolutional Autoencoding Memory) employs a Convolutional AE to obtain spatial features refined in a memory network \citep{CAE-M} and NCAD (Neural Contextual Anomaly Detection) relies on a TCN to capture temporal dependencies \citep{Carmona2022}. Literature points to a clear shift toward spatio-temporal representations as the most dominant strategy for feature extraction in recent DL approaches for AD.

\begin{figure}[h!]
    \centering
    \begin{subfigure}{0.325\textwidth}
        \centering
        \includegraphics[width=\linewidth]{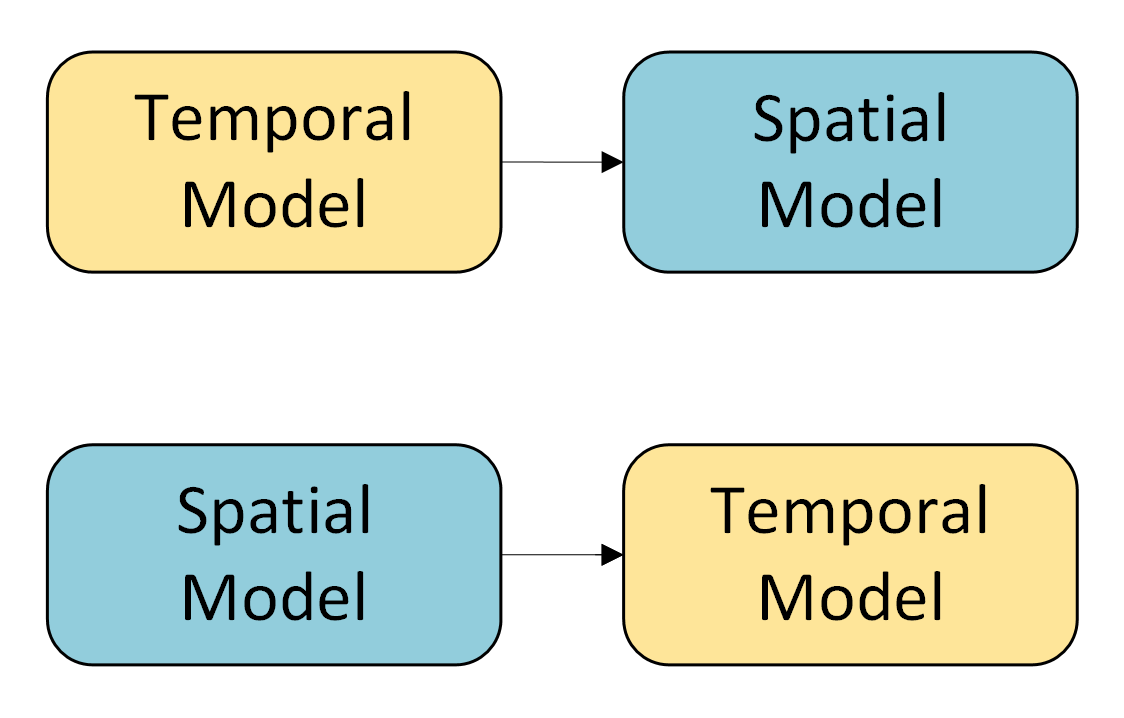}
        \caption{}
        \label{fig:T/S_Seq}
    \end{subfigure}
    \hfill
    \begin{subfigure}{0.325\textwidth}
        \centering
        \includegraphics[width=\linewidth]{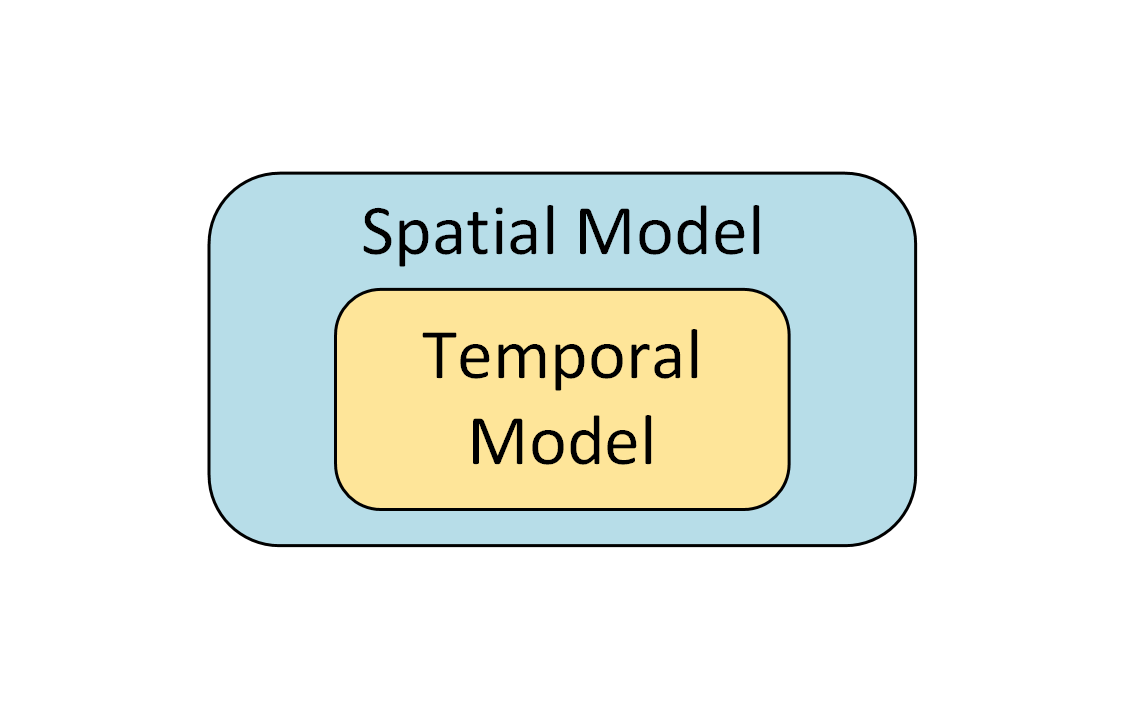}
        \caption{}
        \label{fig:T/S_Nested}
    \end{subfigure}
        \hfill
    \begin{subfigure}{0.325\textwidth}
        \centering
        \includegraphics[width=\linewidth]{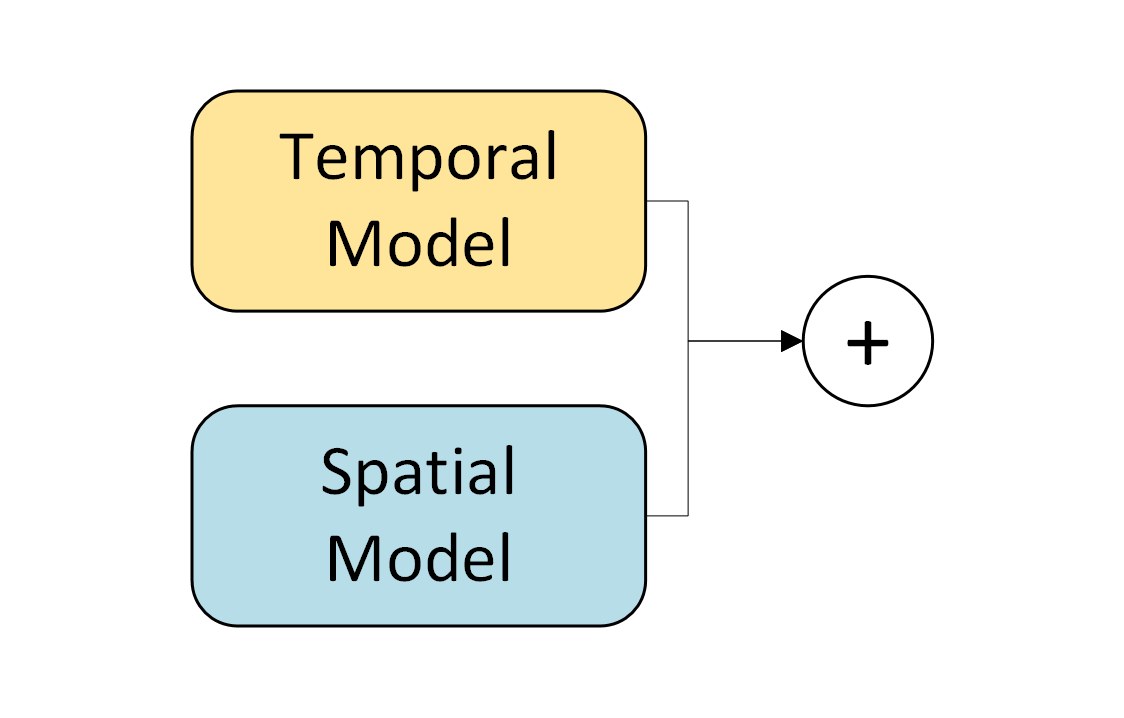}
        \caption{}
        \label{fig:T/S_Parallel}
    \end{subfigure}

    \caption{Schematic diagram representing the novel categories of the T/S dependency dimension, (a) Sequential T$\rightarrow$S (top) and S$\rightarrow$T (bottom), (b) Nested and (c) Parallel.}
    \label{fig: T/S_dependecy}
\end{figure}

Regarding \textbf{T/S dependency} (Temporal~\textbar~Spatial~\textbar~Sequential T$\rightarrow$S~\textbar~Sequential S$\rightarrow$T~\textbar~Nested~\textbar~Parallel~\textbar~Others), spatio-temporal models are more represented than One-dimensional models (either \textit{Temporal} or \textit{Spatial}), with \textit{Nested} and \textit{Other} configurations being more prevalent than \textit{Sequential} and \textit{Parallel} structures. Figure \ref{fig: T/S_dependecy} presents the novel categories of this dimension. Note that the \textit{Nested} configuration shows the most common arrangement (Temporal embedded in Spatial). However, in practice, any combination of spatial, temporal, or spatio-temporal models can be nested within one another. Moreover, the \textit{Other} category is not represented as it covers a variety of different configurations.

To further illustrate the distinction between Feature Extraction and T/S Dependency, consider the following examples. CAE-M is categorized as \textit{Sequential S$\rightarrow$T} in T/S Dependency, but as \textit{Spatial} in Feature Extraction, since it uses a dedicated module only for spatial feature extraction \citep{CAE-M}. Furthermore, GDN is categorized as \textit{Spatial} in both dimensions as it only models spatial dependencies and employs a dedicated GNN-based module to extract spatial features, followed by fully connected layers \citep{GDN}. Finally, despite modeling both spatial and temporal dependencies in a \textit{Nested} configuration, OmniAnomaly lacks a dedicated feature-extraction module and is therefore categorized as \textit{No} in Feature Extraction \citep{OmniAnomaly}. These examples highlight that a model can yield the same category in both dimensions, but this is not necessarily the case. Other examples will be presented in Table \ref{tab: classification}. The following sections explore how different DL architectures are structured to capture temporal and spatial dependencies according to this T/S classification.

\subsubsection{Temporal Models}

Temporal models usually use Transformers. The Transformer model is an encoder-decoder structure, consisting of a stack of multi-head self-attention mechanisms and feed-forward fully connected layers. The multi-head mechanism of the Transformer allows this model to capture various relationships within the data, making it especially effective in MTSAD \citep{Transformer,TranAD}. As an example, TranAD (Transformer-based Anomaly Detection) uses adversarial learning and self-conditioning to train a model consisting of two Transformer encoders and two decoders \citep{TranAD}. 
In contrast, some models rely solely on attention mechanisms without using an encoder-decoder structure. For instance, DCdetector adopts a dual-attention contrastive framework that captures both intra- and inter-patch dependencies, learning discriminative temporal representations directly from the input time window \citep{DCdetector}.

Despite the popularity of Transformer-based models, RNN and TCN are also used within this category \citep{Matar2023,TFAD}.
RNN is a broader term that includes traditional RNN, Long-Short Term Memory (LSTM) networks and Gated Recurrent Units (GRU). Both LSTM and GRU 
effectively capture long-term dependencies, achieving similar performances in many applications. However, GRU has a simpler structure and fewer parameters, leading to faster training times \citep{LSTMandGRU}. The Bidirectional RNN (Bi-RNN), including Bi-LSTM and Bi-GRU, uses two hidden layers to capture both past and future context, improving sequence understanding \citep{BiRNNs}.
The TCN is a CNN-based model, which uses causal and dilated one-dimensional convolutions, to capture long-term dependencies in sequential data \citep{TCN}. As an example, \citet{TFAD} proposed TFAD, which uses the TCN architecture in the time and frequency branches to extract meaningful information in both domains \citep{TFAD}.

\subsubsection{Spatial Models} 

Spatial models focus solely on spatial dependencies and do not explicitly capture temporal dependencies. Typically, MTSAD methods using spatial architectures rely on the sliding window technique during preprocessing to retain temporal information, thus ensuring that the input of the model remains in sequential order \citep{USAD,GDN}. Spatial dependencies can be handled explicitly (e.g. via CNN and GNN) or implicitly (e.g. via AE or GAN).
On the explicit methods, CNN can extract meaningful features from MTS by using a sequence of convolutional layers \citep{MSCRED}. GNN can be used to explicitly learn a graph where each node is a sensor (or dimension) and each edge represents the dependency relationship between two sensors \citep{GDN}. Additionally, GNN can be combined with CNN into Graph Convolutional Networks (GCN) to model node features by aggregating neighbors' features, or with attention mechanisms into Graph Attention Networks (GAT), assigning varying weights to different neighbors, and enhancing the learning of spatial correlation \citep{GCN,GAT}. As an example, GDN learns a graph representing sensor dependencies and applies an attention mechanism to assess their importance, enabling it to predict future sensor values. The prediction process involves an attention function followed by stacked fully connected layers \citep{GDN}.

On the other hand, AE and GAN implicitly capture spatial dependencies through dimensionality reduction.
AE map the input into a lower-dimensional space and decode it back to the original space through an encoder-decoder architecture, learning meaningful information and data characteristics \citep{AE}. Two popular variations of AE widely used in MTSAD are Variational AE (VAE) and Convolutional AE (ConvAE) \citep{OmniAnomaly,MSCRED}. The first introduces probabilistic reasoning into the AE architecture, while the latter incorporates convolutional layers instead of fully connected layers in both the encoder and decoder. GAN train a Generator and the Discriminator through a two-player minimax game \citep{GAN}. This architecture introduced the concept of adversarial training, which is widely employed in MTSAD, as it can help detect anomalies with small deviations from normal by increasing the reconstruction error of encoder-decoder structures \citep{USAD,DAEMON,TranAD}. For example, the USAD model uses adversarial learning to train an AE architecture consisting of an encoder and two decoders. The AE architecture effectively captures spatial dependencies in MTS data, while the adversarial training mechanism amplifies reconstruction errors for anomalous inputs. This approach enables USAD to overcome 
the difficulty of traditional AE in detecting subtle anomalies and the training instability common in GAN \citep{USAD}.

\subsubsection{Sequential Models (T$\rightarrow$S and S$\rightarrow$T)}

Sequential models (Fig. \ref{fig:T/S_Seq}) occupy a small portion of the spatio-temporal models.
An example of a \textit{T$\rightarrow$S} model was presented by \citet{Choi2020}, which developed a GAN-based model and employed a point-wise convolutional layer on top of the generator to extract the temporal dependency. 
Within the \textit{S$\rightarrow$T} category, \citet{Tang2023} proposed the spatial GDN model before the temporal GRU prediction after the attention mechanism. This model first learns the relationships between sensors through GDN and then utilizes a GRU that models temporal information \citep{Tang2023}. As another example, CAE-M uses a ConvAE for deep feature extraction and then concatenates the latent representation to the reconstruction error, and feeds them into a memory network for prediction. This network performs non-linear prediction through Bi-LSTM followed by an attention mechanism and combines it with linear prediction using an autoregressive (AR) model \citep{CAE-M}. Additionally, MST-GAT (Multimodal Spatial-Temporal GAT) employs a multimodal GAT, with three attention modules (multi-head attention, intra-modal attention and inter-modal attention) to capture spatial and multimodal correlations across different dimensions. The outputs from these attention modules are then concatenated and passed through a TCN to effectively model temporal dependencies \citep{MST-GAT}.

\subsubsection{Nested Models}

In Nested models (Fig. \ref{fig:T/S_Nested}), typically, a temporal model is embedded within an AE a GAN-based model. 
For instance, several studies have implemented LSTM-AE architectures, where the encoder and decoder structure of an AE is composed of LSTM layers \citep{Kieu2019,Nguyen2021}.
As another example, the OmniAnomaly uses GRU incorporated into a VAE framework to create a stochastic RNN to learn latent representations and capture temporal dependencies \citep{OmniAnomaly}. Additionally, DAEMON (adversarial Autoencoder anomaly detection interpretation) uses adversarial training in an AE-based model, where encoder and decoder networks consist of 1D-CNN, which apply filters that slide along the temporal dimension, to capture the temporal dependencies \citep{DAEMON}. 
An example of a GAN-based model is MAD-GAN (Multivariate Anomaly Detection with GAN), which employs an LSTM network as the base model for both the Generator and Discriminator. MAD-GAN considers the entire dimension set simultaneously in order to capture the latent interactions among the different dimensions and uses LSTM to capture temporal dependency \citep{MAD-GAN}. 

The \textit{Nested} category also comprise spatio-temporal models embedded in a spatial model.
As an example, MSCRED encodes the spatial correlations of signature matrices at different resolutions through a ConvAE and embeds an attention-based ConvLSTM in its architecture to model temporal information \citep{MSCRED}. The ConvLSTM is an spatio-temporal model that uses convolutional operations inside the LSTM cells instead of the traditional fully connected layers \citep{convLSTM}. MSCRED adds an attention mechanism to this model to adaptively select relevant hidden states \citep{MSCRED}. 
Some variations of MSCRED have been proposed \citep{Xie2023,Yokkampon2022}. \citet{Yokkampon2022} implemented a convolutional VAE as the backbone structure instead of the convolutional encoder-decoder used in MSCRED. Furthermore, \citet{Xie2023} added a VAE before the convolutional decoder, to learn the distribution of the high-level representations of the outputs from ConvLSTM.
Using a different DL architecture, \citet{Yu2021} proposed a GAN-based model, where each network was composed of a CNN-LSTM architecture. Unlike ConvLSTM used in the MSCRED model, the CNN-LSTM architecture employs a LSTM network after the convolutional layers. In this case, a time-scale and a spatial-scale convolutional layer are concatenated to extract temporal and spatial features, respectively, before the LSTM further captures temporal dependencies. In another approach, \citet{Lian2023} proposed a model where \textit{S$\rightarrow$T} structure is nested in a GAN architecture. Both the Generator and the Discriminator are composed of a GAT layer to extract spatial correlation followed by a Bi-LSTM and an attention mechanism to capture the temporal dependencies.

\subsubsection{Parallel Models} Parallel models (Fig. \ref{fig:T/S_Parallel}) handle both dependencies simultaneously using a two-branch architecture, one dedicated to temporal dependencies and the other to spatial dependencies. Typically, a temporal model is employed in one branch and a spatial model in the other. For example, the STADN (AD Network using Spatial and Temporal information) models spatial dependencies using a GAT and temporal dependencies using an LSTM. The two feature representations are then concatenated and passed through stacked fully connected layers for prediction  \citep{Tian2023}. Similarly, ESTAD (Enhanced Spatio-Temporal constraints end-to-end network for AD) employs a Transformer encoder to extract temporal features in parallel with a GNN model that learns spatial dependencies. It further introduces constraint tasks in both temporal and spatial domains, such as an autoregressive temporal objective and graph contrastive learning, to enhance spatio-temporal representations \citep{ESTAD}. 

Parallel models may also use a mirrored architecture, where both branches share the same structure but are adjusted for their respective dependencies. For example, PCAC (Parallel Convolutional Anomaly multi-Classification) employs two parallel sub-networks, each composed of a 1D-CNN architecture followed by an attention mechanism. These branches extract spatial and temporal features in parallel, and the resulting representations are concatenated before classification \citep{PCAC}.
Furthermore, BTAD (Bi-Transformer Anomaly Detection) employs a Bi-Transformer architecture to model the two dimensions in parallel, incorporating an enhanced adaptive multi-head attention mechanism and a modified decoder structure \citep{Ma2023}.

\subsubsection{Other Models}

This category includes spatio-temporal models that do not conform to the aforementioned frameworks, featuring diverse architectures with no consistent organization.
As an example, GTA links graph convolutions with hierarchical dilated convolutions on the original MTS to capture both temporal and spatial dependencies \citep{GTA}. Then, the output of this context block serves as input to a Transformer to further capture temporal dependencies.
Additionally, STGAT-MAD (Spatial-Temporal GAT for Multivariate time series AD) uses stackable GAT to capture both feature and temporal correlations across different time scales \citep{Zhan2022}.
In a separate study, VGCRN (Variational Graph Convolutional Recurrent Network) integrates multiple GCRN modules, combining GCN with a recurrent structure to capture spatial and temporal dependencies. By stacking these modules, VGCRN enhances the ability to model multilevel temporal dependencies within MTS data. This architecture is combined with a deep embedding-guided probabilistic network to enhance generalization capacity with hierarchical prior and capture the dependencies between channels in embedding space \citep{DVGCRN}.

Within the \textit{Others} category, some models combine the structures previously described in an hybrid fashion. For instance, a group of models employs a \textit{Parallel} module followed by a \textit{Temporal} model to capture long-term patterns, resembling a \textit{Sequential S$\rightarrow$T} architecture in which the spatial module is replaced by a parallel block.
As an example,
the MTAD-GAT first employs a 1D-CNN layer to extract high-level features of each TS. Then, the output is processed by two parallel GAT, one time-oriented and the other feature-oriented.
The resulting representations are concatenated with the convolution output and fed into a GRU to model long-term dependencies \citep{MTAD-GAT}. 
Several adaptations of this model have been proposed \citep{Song2023,Zhou2022,Xiong2023}. In particular, \citet{Zhou2022} removed the 1D-CNN layer, whereas \citet{Xiong2023} replaced it with a Transformer encoder. Finally, \citet{Song2023} introduced a multi-head self-attention mechanism before the GRU, alongside a first-order AR model running in parallel to enhance linear prediction.

The dominance of spatio-temporal models reflects the inherent complexity of MTS data, where anomalies often manifest in both temporal and spatial dependencies simultaneously. \textit{Temporal} models are a natural choice when spatial dependencies are weak, while \textit{Spatial} models are more appropriate when anomalies arise from disruptions in inter-sensor relationships. Among spatio-temporal configurations, \textit{Sequential} models are appropriate when one dependency type is less critical compared to the other, \textit{Parallel} models when both carry equal importance, and \textit{Nested} models when temporal and spatial dependencies are tightly coupled and best captured within a unified architecture. Finally, \textit{Other} models represent architectures that combine elements from multiple configurations in hybrid or unconventional ways. At this stage, the proposed taxonomy already provides a structured foundation that can be naturally extended as new architectural patterns consolidate in the literature.

\section{Past, present and future research directions}\label{sec5}

The proposed taxonomy is the result of an in-depth analysis of more than 400 papers, manually selected from a set of documents retrieved from Scopus using a reproducible search query. Although the search term ``multivariate time series'' may appear somewhat restrictive, its use ensured the inclusion of relevant contributions while excluding studies that did not explicitly address multivariate settings. The research field of MTSAD is highly active, with several reviews already published in 2025 \citep{Wang2025, Paparrizos2025, Jia2025}. For instance, \citet{Jia2025} present a classification of more than 200 papers based on loss functions. While this perspective is valuable, it does not include the broader methodological dimensions of DL-based MTSAD. The taxonomy introduced in this work thus complements existing surveys and is designed to remain adaptable, allowing new categories or dimensions to be integrated as the field advances.

\subsection{Overview until 2024}

Before 2019, MTSAD research was dominated by traditional ML methods, such as clustering, Support Vector Machines and tree-based models, which were valued for their interpretability and their ability to capture anomaly patterns across multiple variables. Although such works were identified in the survey, they were not included in the review since the focus lies on DL-based approaches. Readers interested in ML-based strategies are referred to existing surveys \citep{BlzquezGarca2021, Paparrizos2025}. From 2019 onwards, a clear shift towards DL-based models can be observed. The first wave of contributions, particularly between 2019 and 2022, introduced disruptive architectural innovations, such as OmniAnomaly \citep{OmniAnomaly}, MAD-GAN \citep{MAD-GAN}, MSCRED \citep{MSCRED}, USAD \citep{USAD}, MTAD-GAT \citep{MTAD-GAT}, GDN \citep{GDN}, TranAD \citep{TranAD}, GTA \citep{GTA} and Anomaly Transformer \citep{AnomalyTransformer}. More recent works, published between 2023 and 2024, tend to focus on incremental improvements, either by fine-tuning existing architectures, modifying loss functions, or refining input and output representations. Nevertheless, notable exceptions remain, with models such as CAE-M \citep{CAE-M} and DCdetector \citep{DCdetector} introducing novel architectural concepts after 2022. Table \ref{tab: classification} presents the categorization of these representative methodologies across the
dimensions of the proposed taxonomy, illustrating their different architectural and methodological choices and enabling a structured comparison of the approaches within the taxonomy.

\begin{sidewaystable}
\centering
\caption{Classification of a set of representative MTSAD methodologies on the dimensions of the proposed taxonomy.} 
\label{tab: classification}
\footnotesize
\begin{tabular}{ccccccccccccc}
\toprule
{\textbf{Part}} & {\textbf{Dimension}} & 
\begin{tabular}[c]{@{}c@{}}OmniAnomaly\\(2019)\end{tabular} & 
\begin{tabular}[c]{@{}c@{}}MAD-GAN\\(2019)\end{tabular} & 
\begin{tabular}[c]{@{}c@{}}MSCRED\\(2019)\end{tabular} & 
\begin{tabular}[c]{@{}c@{}}USAD\\(2020)\end{tabular} & 
\begin{tabular}[c]{@{}c@{}}MTAD-GAT\\(2020)\end{tabular} & 
\begin{tabular}[c]{@{}c@{}}GDN\\(2021)\end{tabular} & 
\begin{tabular}[c]{@{}c@{}}TranAD\\(2022)\end{tabular} & 
\begin{tabular}[c]{@{}c@{}}GTA\\(2022)\end{tabular} & 
\begin{tabular}[c]{@{}c@{}}Anomaly\\Transformer\\(2022)\end{tabular} & 
\begin{tabular}[c]{@{}c@{}}CAE-M\\(2023)\end{tabular} & 
\begin{tabular}[c]{@{}c@{}}DCdetector\\(2023)\end{tabular} \\
\midrule
\multirow{3}{*}{\textbf{Input}}  & 
\textbf{Preprocessing} & N & N & Y & N & N & N & N & N & N & N & N \\ 
\cmidrule{2-13}
& \textbf{Format} & W & W & I & W  & W  & W & W  & W & W & W & W \\ 
\cmidrule{2-13} 
& \textbf{Content} & Time & Time & Others & Time & Time & Time & Time & Time & Time & Time & Time \\ 
\midrule
\multirow{6}{*}{\textbf{Output}} & \textbf{Type} & Rec & Rec & Rec & Rec & Hybrid   & Pred & Rec & Pred & Rec & Pred & Rep \\ 
\cmidrule{2-13} 
& \textbf{\begin{tabular}[c]{@{}c@{}}Optimal\\ Threshold\end{tabular}}  & Y & n.e. & Y & Y & Y & Y & Y & Y & Y & Y & Y \\ 
\cmidrule{2-13} 
& \textbf{\begin{tabular}[c]{@{}c@{}}Time-variyng\\ Threshold\end{tabular}} & N & N & N & N & N & N & Y & N & N & N & N \\ \cmidrule{2-13} 
& \textbf{Granularity} & Point & Point & Subseq & Point & Point & Point & Point & Point & Point & Subseq & Point \\ 
\midrule
\multirow{7}{*}{\textbf{Model}}  & \textbf{Task} & Reg & Reg & Reg & Reg & Reg & Reg & Reg  & Reg & Reg & Reg & Reg \\
\cmidrule{2-13} 
& \textbf{\begin{tabular}[c]{@{}c@{}}Loss\\ Function\end{tabular}} & ELBO & Adv & SumE & SumE & Multi & SumE & SumE & SumE & Multi & Multi & Con \\ 
\cmidrule{2-13} 
& \textbf{\begin{tabular}[c]{@{}c@{}}Feature\\Extraction\end{tabular}} & N & N & N & N & ST & S & N & ST & N & S & T \\ 
\cmidrule{2-13} 
& \textbf{\begin{tabular}[c]{@{}c@{}}T/S\\ Dependency\end{tabular}}   & Nested & Nested & Nested & S & Others & S & T & Others & Others & S$\rightarrow$T & T \\ 
\bottomrule 
\end{tabular}
\begin{minipage}{\linewidth}
{\footnotesize  N: No; Y: Yes; W: Window; I: Image; Rec: Reconstruction; Pred: Prediction; Rep: Representation; Subseq: Subsequence; Reg: Regression; Adv: Adversarial; Multi: Multi-objective; Con: Contrastive; T: Temporal; S: Spatial; ST: Spatio-Temporal}
\end{minipage}
\end{sidewaystable}

The categorization of 409 articles highlights several trends in the field for 2019--2024. Most studies rely on raw time series organized into sliding windows while preserving the temporal order, although there is a growing tendency to represent time series as images to capture spatial correlations. Regression-based settings dominate, and distance-based losses, such as MSE, remain the most widely adopted due to their simplicity, even though hybrid loss functions have been increasingly explored to capture more complex patterns. 
In temporal/spatial dependency modeling, sequential, nested, and other complex structures prevail, with nested and heterogeneous architectures being particularly common as these approaches jointly capture temporal and spatial dependencies. 
Moreover, parallel structures are gaining visibility due to their scalable nature, enabling simultaneous modeling of multiple dependencies and improving computational efficiency.
Dedicated feature extraction modules are relatively rare, but when present, they typically focus on extracting spatio-temporal features. With respect to the output, anomaly scores are most often computed through reconstruction errors, although hybrid strategies that combine reconstruction errors with prediction errors or latent space deviations are gaining popularity. Detection continues to be primarily carried out at the point level, with thresholds that are generally optimal but time-invariant. The choice of architecture has also evolved, with Transformers and GNN becoming increasingly prominent. Transformers are particularly effective for temporal modeling, while GNN are widely adopted to capture spatial dependencies. More recent directions include the exploration of generative models, such as Normalizing Flows and Diffusion models, as well as contrastive learning strategies. The application of Large Language Models (LLM) to MTSAD is still at an early stage, but their adoption is increasing, reflecting the broader impact of Transformer-based architectures in time series research \citep{GPT4TS, MOMENT, DualLMAD}.

The dimensions of the proposed taxonomy are independent, and each article is classified into a single category per dimension. Nevertheless, certain patterns may emerge across dimensions. The most frequent combination includes \textit{No} Preprocessing, \textit{Window} Format, \textit{Time} Content, \textit{Reconstruction} Type, an \textit{Optimal} but \textit{Time-invariant} Threshold, \textit{Point} Granularity, \textit{Regression} Task, \textit{ELBO} Loss Function, \textit{Nested} T/S Dependency, and \textit{No} Feature Extraction module. This profile is largely driven by the popularity of LSTM-VAE \citep{LSTM-VAE}, one of the earliest DL approaches for MTSAD, which established reconstruction-based AD with uncertainty estimation as a dominant paradigm, with subsequent works building upon and extending its core methodology, as exemplified by OmniAnomaly \citep{OmniAnomaly}, which further incorporated GRU and NF to improve latent space modeling.
However, this setup occurs in only 18 out of the 409 categorized papers, highlighting the heterogeneity of approaches. Since the Loss Function, Optimal Threshold and Time-varying Threshold dimensions highly contributed to this variability, they were ignored for subsequent analyses. When ignoring these three dimensions, the number of papers matching this combination rises to 79 out of 409 (i.e. 19.32\%). 
This joint proportion is larger than the product of proportions over each dimension, which correponds to the case of categories being chosen independently ($\sim$5\%). This indicates a positive association such that that the presence of a category in one dimension increases the likelihood of a category in the remaining dimensions.

\begin{figure}[h!]
    \centering
    \includegraphics[width=\linewidth]{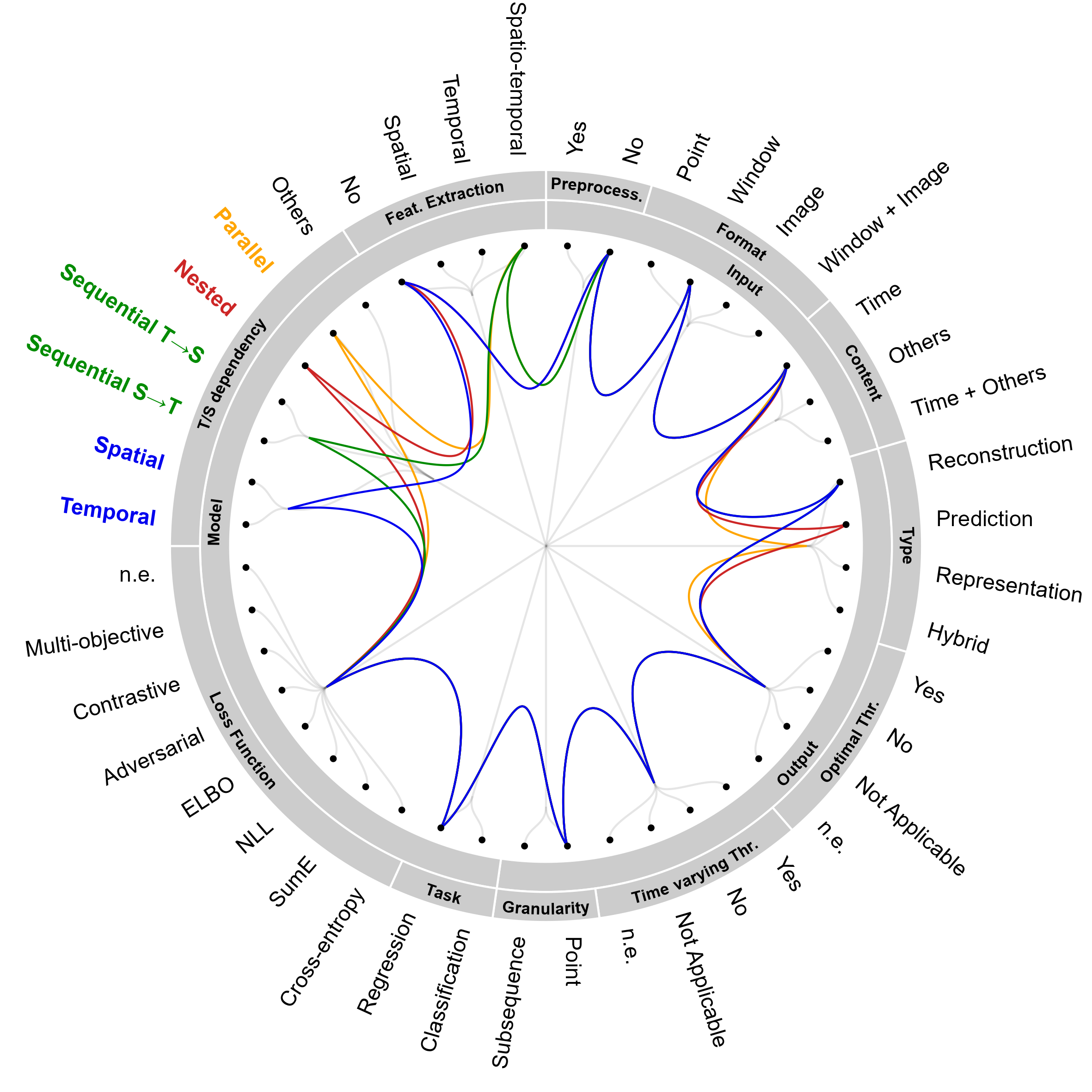}
    \caption{Configuration patterns of models across T/S dependency categories (One-dimensional, Sequential, Nested and Parallel). Analysis excluded the Loss Function, Optimal Threshold, and Time-varying Threshold dimensions.}
    \label{fig:bundling}
\end{figure}

Figure \ref{fig:bundling} shows the same analysis conditional on the T/S dependency categories, namely One-dimensional, Sequential, Nested and Parallel models. The One-dimensional (Temporal and Spatial) and Sequential (T$\rightarrow$S and S$\rightarrow$T) models were grouped, as they display similar patterns. The Others category was excluded due to its high internal variability. One-dimensional and Nested models tend to rely on \textit{Reconstruction} errors without a Feature Extraction module, whereas Sequential models more often combine \textit{Prediction} errors with dedicated \textit{Spatio-Temporal} feature extraction. In all three cases, the joint proportion (One-dimensional: 27.88\%; Sequential: 29.51\%; Nested: 63.71\%) exceeds the proportion expected under independence (One-dimensional: 17.21\%; Sequential: 15.26\%; Nested: 46.61\%), showing that choices across dimensions are not independent.
In contrast, Parallel structures typically include a \textit{Spatio-Temporal} module for feature extraction but exhibit no consistent pattern in Type. Within this configuration, all Type categories have the same joint proportion of 13.89\%. Thus, removing the Type dimension, the joint proportion rises to 55.26\%, compared to 46.18\% expected under independence. 
Overall, this analysis indicates that architectural design choices in MTSAD models are interdependent, forming coherent and consistent patterns that vary according to T/S dependency. However, unlike other dependency categories, Parallel structures lack a consistent pattern, suggesting a greater variability in how the categories are combined. Finally, it is worth noting that \textit{n.e.} (not explicit) entries do not reflect a limitation in taxonomy coverage but rather a reporting limitation in the categorized literature. This is particularly concerning for the Thresholding and Loss Function dimensions, which are critical components of any MTSAD approach. The lack of explicit reporting on these aspects compromises transparency and reproducibility in the research field.

The proposed taxonomy is suitable to classify emerging paradigms applied to MTSAD, including foundation models such as TimesNet \citep{TimesNet}, LLM-based approaches \citep{DualLMAD}, and distributed approaches, such as federated learning-based methods \citep{FDVL-DCN}. These models still rely on the same core design decisions captured by the taxonomy dimensions. As an example, DualLMAD \citep{DualLMAD} employs two pre-trained GPT-2 models in a \textit{parallel} fashion to simultaneously capture T/S dependencies within the time series. The model reconstructs the input \textit{window} and computes the anomaly score as \textit{reconstruction} error, using an \textit{optimal} and \textit{time-varying} threshold to obtain the final classification. While these approaches differ in their training paradigm, whether through large-scale pre-training or federated aggregation, their underlying detection methodology remains fully compatible with the proposed framework, confirming that it can accommodate future developments in MTSAD.

\subsection{Some developments in 2025}

A search of publications from 2025 (up to 31 October) using the same criteria as the original taxonomy yielded 320 papers, with 190 selected for detailed examination.
The analysis of these papers highlights a research field marked by the stabilization of the categories of the proposed taxonomy and the introduction of hybrid or extensions of the previous formulations. 

\textbf{Input}-related dimensions (Pre-processing, Format and Content) show continuity with earlier years with some broadening. Most recent contributions continue to rely on \textit{Window}-based formats for modeling subsequences. As examples, \citet{MLAD} adopt fixed-length sliding windows to capture localized temporal dependencies, while \citet{TRLP} introduce multi-resolution pyramid windows that model dynamics at multiple temporal scales. Together, these papers exemplify how contemporary research refines rather than replaces traditional window-based representations in MTSAD. At the same time, several studies move beyond purely temporal segmentation by integrating \textit{Window+Image} representations or frequency-transformed inputs. For instance, \citet{FDTAD} incorporate frequency-domain features alongside temporal windows, and \citet{FreqWave-TranDuD} make use of wavelet-based frequency transformations to enhance multiscale sensitivity. These trends consolidate the \textit{Preprocessing} Input dimension within the taxonomy and highlight a broader shift toward hybrid representations that merge temporal, spatial and frequency-domain information. Window-based formats remain dominant as they are compatible with most DL architectures and support the detection of Point, Contextual and Collective anomalies. The move toward hybrid content reflects the understanding that temporal information alone may be insufficient to capture all anomaly types, with frequency and spatial information providing complementary signals that improve detection.

At the \textbf{Output} level, the categories of \textit{Reconstruction} and \textit{Prediction} remain central, but 2025 research expands the use of \textit{Representation}-based deviations and \textit{Hybrid} anomaly scores. AD2T (Association Discrepancy Dual-decoder Transformer) combines reconstruction, prediction and association discrepancy in a dual-decoder framework \citep{AD2T}, while TRLP (Tranformer-based Reconstruction-Limited Probability) introduces probabilistic scoring based on representation uncertainty \citep{TRLP}. Furthermore, adaptive thresholding mechanisms gained traction e.g. Mixer-transformer employs time-varying thresholds to reduce false positives \citep{Mixer-transformer}, while boundary-aware ensembles generate thresholds based on estimated uncertainty \citep{He2025}. 
Together, these papers reinforce that \textit{Hybrid} type outputs are gaining ground as they exploit the complementary strengths of reconstruction, prediction and representation-based signals, improving the detection of diverse and complex anomaly patterns. Moreover, \textit{time-varying thresholds} allow models to adapt to the non-stationary nature of real-world data.
For \textit{Granularity}, most contributions still detect anomalies at the \textit{Point} level. For example, methodologies such as MLAD (Multi-task Learning Anomaly Detection) and MGDRF (Multi-Grain Dynamic Receptive Field) explicitly model multi-scale subsequences, yet they assign anomaly scores to each time point by considering its temporal context \citep{MLAD, MDGRF}. These developments indicate that point granularity anomaly detection remains a natural choice for most DL-based MTSAD methodologies in 2025, as it provides the finest temporal resolution for anomaly localization and is directly compatible with window-based input formats.

The \textbf{Model} part is where 2025 contributions show the greatest methodological innovation. In the \textit{Task} dimension, explicit \textit{Classification} formulations remain uncommon and \textit{Regression} continues to dominate, with models learning to identify deviations as anomalies \citep{MLAD, TRLP, DiffANT, TiTAD, TSAD}.
Moreover, several 2025 approaches integrate prediction with other representation learning (e.g. contrastive, variational or adversarial) to enhance the discriminative capacity of latent features \citep{RGAnomaly, DyGCL, IGANomaly,AVAE}. As a consequence, the \textit{Loss Function} dimension shows some relevant developments. Beyond the classical reconstruction losses (\textit{SumE}, \textit{ELBO}), the 2025 research incorporates more frequent \textit{Contrastive} objectives (CARLA \citep{CARLA}, DMAP-DDCL \citep{DMAP-DDCL}), \textit{Adversarial} losses (EH-GAM-EGAN \citep{EH-GAM-EGAN}, Quantum GAN \citep{Quantum_GAN}) and multi-objective criteria that merge reconstruction, prediction and representation learning (AD2T \citep{AD2T}, STAMP \citep{STAMP}). These advances consolidate the\textit{ Multi-objective} category as a mainstream choice for the \textit{Loss Function} dimension. Overall, this reflects a progression from simple error minimization toward richer training objectives, allowing models to capture more discriminative representations of normal behavior and thus improve sensitivity to subtle and complex anomalies.

The \textit{T/S Dependency} in 2025 shows an evolution beyond the classical \textit{Temporal}-only or \textit{Sequential T$\rightarrow$S} paradigms. Recent architectures increasingly emphasize explicit and bidirectional modeling of temporal and spatial interactions. For instance, MSTGAD (MTSAD based on multiple spatiotemporal Graph convolution) introduces DTW-based spatio-temporal graphs, aligning temporal similarity with dynamic spatial adjacency to enhance relational reasoning \citep{MSTGAD}, while HiGraph (Hierarchical Graph) captures hierarchical dependencies across intra- and inter-series levels, merging local and global context in a unified graph hierarchy \citep{HiGraph}. 
Similarly, HYMAN (HYbrid Memory and Attention Network) explicitly separates local temporal attention from global memory aggregation \citep{HYMAN}. 
Collectively, these designs reinforce the emergence of \textit{Nested}, \textit{Parallel} and hybrid dependency categories, marking a shift toward architectures that treat temporal and spatial structures as co-evolving rather than just sequentially coupled, allowing the detection of anomalies that span both time and variable relationships that would otherwise be missed.
Finally, 2025 research demonstrates clear progress toward \textit{Spatio-Temporal} feature extraction modules. Examples include the VGATSL (Variational Graph ATtention networks with Self-supervised Learning) that couples variational graph neural networks with self-supervised objectives \citep{VGATSL} and the MemMambaAD, which adopts a state-space model with adaptive memory dynamics \citep{MemMambaAD}. This suggests that explicitly encoding spatio-temporal features before anomaly estimation may lead to better representations of normal behavior, making anomalies easier to detect than when these representations are learned implicitly by the MTSAD model.

\subsection{Future research directions and open challenges}

Building on the analysis presented throughout this review, several open challenges and directions emerge that are likely to shape the future of MTSAD research. In particular, three main directions can be identified
\begin{itemize}
\item Adaptive, interpretable and scalable spatio-temporal architectures;
\item Moving beyond exclusively DL-based approaches for MTSAD;
\item Awareness of the limitations of existing benchmarks. 
\end{itemize}
Each of these directions is discussed below.

The first direction concerns the development of adaptive, interpretable and scalable spatio-temporal architectures for MTSAD.
Graph-Transformer hybrids such as LGAT exemplify the value of unifying attention-based temporal modeling with dynamic graph reasoning for non-stationary data \citep{LGAT}. Generative models like MAFCD (Multi-level and Adaptive Conditional Diffusion) emphasize multi-view probabilistic learning for robust anomaly synthesis \citep{MAFCD}, while memory-augmented designs such as MemMambaAD enhance long-term contextual memory \citep{MemMambaAD}. Additionally, explainable models like SHAPAttenVAE (Shapley Attention VAE) signal growing attention to interpretability and causal reasoning in the field \citep{SHAPAttenVAE}. Real-world deployment demands models that go beyond accuracy, requiring transparency, adaptability to non-stationary environments and computational efficiency. Therefore, these directions are likely to become central to the next generation of MTSAD research. 

The second direction reflects a growing recognition of the need to move beyond exclusively DL-based approaches for MTSAD. While DL-based methods currently dominate the field, several studies question whether this paradigm consistently represents the most effective solution across diverse settings \citep{Audibert2022, Wu2021}. Although DL methods tend to handle large datasets more efficiently than conventional or ML-based approaches, there is no conclusive evidence that the first consistently outperform the latter across a wide range of datasets and real-world conditions. Moreover, reproducibility remains a major obstacle for adoption in real scenarios, as multiple factors, including model design, hyperparameter settings and computational resources influence results. These challenges suggest that the field should not rely exclusively on DL-based paradigms. In certain applications, classical approaches, ML-based methods or hybrid frameworks may offer more suitable and interpretable solutions. Furthermore, integrating domain knowledge through expert systems into data-driven models represents a promising direction, combining the strengths of artificial intelligence with human expertise to improve robustness, reliability and applicability in real-world anomaly detection tasks. Rethinking the DL-only paradigm is critical, as increasing model complexity does not guarantee better performance in practice and may further distance research from real-world applicability.

The third direction relates to the limitations of current benchmark datasets and evaluation protocols. Although this review does not provide a detailed analysis of benchmark datasets, as this topic has been extensively covered in prior work \citep{ZamanzadehDarban2024, Garg2022}, it is important to acknowledge ongoing concerns regarding their reliability and representativeness. Nonetheless, concerns remain about the reliability of widely used datasets such as MSL \citep{MSL}, SMAP \citep{MSL} and SMD \citep{OmniAnomaly}. \citet{Wu2021} argue that these benchmarks may \textit{create an illusion of progress} due to trivial anomalies, mislabeled instances and unrealistic conditions. Without reliable and diverse benchmarks, it becomes difficult to determine whether improvements reflect genuine advances or simply overfitting to the particularities of widely used datasets. Therefore, the development of new, high-quality and diverse datasets is essential to enable fair comparisons between approaches and genuine quantification of progress in the field.

\section{Conclusions}\label{sec6}

This paper presents a thorough and detailed systematic literature review of DL–based methods for MTSAD, offering a comprehensive perspective of a rapidly expanding research field.
A novel and unified taxonomy is introduced, encompassing eleven dimensions grouped into three main parts: Input, Model, and Output. Together, these dimensions provide a structured and unified way to compare and classify DL-based methods for MTSAD. Designed with sufficient detail to systematically distinguish among different approaches, yet flexible enough to accommodate future developments, this taxonomy captures the essential characteristics that define the diversity of methods in the field.

This comprehensive tool was developed based on the analysis of methodological papers and on the insights of previous review papers. The categorization of 409 papers offers a snapshot of the current research landscape, highlighting dominant methodologies and emerging paradigms. Furthermore, the taxonomy was validated on an additional set of papers published 2025, confirming its relevance and adaptability to ongoing developments in the field. Moreover, an analysis of the associations between dimensions showed a tendency for specific model architectures to align with particular output types, highlighting that methodological choices are often interdependent, with certain categories frequently co-occurring.

This paper presents a framework that enables the development of novel methods, supports informed methodology choices, and guides future research toward addressing open challenges in MTSAD. By analyzing current trends, identifying underlying patterns, and highlighting emerging directions, it offers researchers and practitioners a practical tool to understand and navigate the field.

\backmatter

\bmhead{Supplementary information}
The supplementary material includes a full list of selected documents and reviews analysed in this paper.

\section*{Declarations}

\subsection*{Funding}
This work was supported by IEETA (\href{https://www.ieeta.pt/}{https://www.ieeta.pt/}, UID/00127)
through the funding of the Portuguese Foundation for Science and Technology 
(FCT, \href{https://www.fct.pt/}{https://www.fct.pt/}, \href{https://ror.org/00snfqn58}{https://ror.org/00snfqn58}), in the scope of the R\&D Grants UID/00127/2025 (\href{https://doi.org/10.54499/UID/00127/2025}{https://doi.org/10.54499/UID/00127/2025})
and UID/PRR/00127/2025. B.A. acknowledges the individual PhD scholarship (ref. 2024.00557.BD), funded by national funds through FCT.

\subsection*{Conflict of interest}
The authors declare no conflict of interest.






\bibliography{references}

\end{document}